\documentclass[a4paper,12pt,twoside]{report}

\usepackage[left=3cm,right=3cm,top=3cm,bottom=3cm]{geometry}

\usepackage{lipsum}
\usepackage{moreverb,url}
\usepackage[bookmarks=true, colorlinks,bookmarks=true,citecolor=blue,urlcolor=blue]{hyperref}
\usepackage{natbib}
\usepackage{epstopdf}
\usepackage{amsmath}
\usepackage{amssymb}
\usepackage{latexsym}
\usepackage{url}
\usepackage{relsize}
\usepackage{multirow}
\usepackage{afterpage}
\usepackage{ifthen}
\usepackage{graphicx}
\usepackage{algpseudocode,algorithm,algorithmicx}
\usepackage{subfigure}
\usepackage[keeplastbox]{flushend}
\usepackage{epstopdf}
\usepackage{stackengine}
\usepackage{gensymb}
\usepackage{booktabs}
\usepackage{makecell}
\usepackage{hhline}
\usepackage{courier}

\usepackage{makecell}
\usepackage[10pt]{moresize}
\usepackage{anyfontsize}

\usepackage[figurename=Fig.]{caption}  % change Figure to Fig.

\hypersetup{pdfstartview={FitH}}

\setcitestyle{authoryear,open={(},close={)}}
\setcitestyle{aysep={,}} 

\newcommand{\ra}[1]{\renewcommand{\arraystretch}{#1}}

\newcommand \tableCaption[6]{ %4 is number of cells
    \begin{tabular}{l@{\hspace{2.0\tabcolsep}}l} %2 columns - "c" for center; "l" for left
    \scriptsize  \textbf{GT}: #1 		&	\scriptsize \textbf{SGC}: #3\\     % #1 IS FOR REAL TEXT VALUE
    \scriptsize  \textbf{S2SNet}: #2 	&	\scriptsize \textbf{S2VT}: #4\\    %2 rows
    \scriptsize  \textbf{V2CNet}: #5 	&	\scriptsize \textbf{SCN}: #6\\    %3 rows
     \end{tabular}
}

\newcommand \datasetCaption[2]{ %4 is number of cells
    \begin{tabular}{ll} %2 columns - "c" for center; "l" for left
         	\scriptsize \textbf{Command}: #1 	&	\scriptsize \textbf{Action}: #2     % #1 IS FOR REAL TEXT VALUE
         	\vspace{1.5ex}
     \end{tabular}
}

\usepackage{color, colortbl}
\definecolor{Gray}{gray}{0.9}

\usepackage{graphicx}
\usepackage{verbatim}
\usepackage{latexsym}
\usepackage{mathchars}
\usepackage{setspace}

\input{blocked.sty}
\input{uhead.sty}
\input{boxit.sty}
\input{icthesis.sty}

\newcommand{\NN}{{\sf I\kern-0.14emN}}   % Natural numbers
\newcommand{\ZZ}{{\sf Z\kern-0.45emZ}}   % Integers
\newcommand{\QQQ}{{\sf C\kern-0.48emQ}}   % Rational numbers
\newcommand{\RR}{{\sf I\kern-0.14emR}}   % Real numbers

\newcommand{\normallinespacing}{\renewcommand{\baselinestretch}{1.5} \normalsize}

\newcommand{\syncc}{~\stackrel{\textstyle \rhd\kern-0.57em\lhd}{\scriptstyle L}~}

\RequirePackage{caption}
\DeclareCaptionLabelSeparator{sageperiod}{.\hspace*{1ex}}
\begin{document}

\title{\LARGE {\bf Scene Understanding for Autonomous Manipulation with Deep Learning}\\
 \vspace*{6mm}
}

\author{Anh Nguyen}
\submitdate{November 2018}

\normallinespacing
\maketitle

\preface
% Macros
\newcommand{\junk}[1]{} %% just for comment

\addcontentsline{toc}{chapter}{Abstract}

\begin{abstract}
Over the past few years, deep learning techniques have achieved tremendous success in many visual understanding tasks such as object detection, image segmentation, and caption generation. Despite this thriving in computer vision and natural language processing, deep learning has not yet shown significant impact in robotics. Due to the gap between theory and application, there are many challenges when applying the results of deep learning to the real robotic systems. In this study, our long-term goal is to bridge the gap between computer vision and robotics by developing visual methods that can be used in real robots. In particular, this work tackles two fundamental visual problems for autonomous robotic manipulation: affordance detection and fine-grained action understanding. Theoretically, we propose different deep architectures to further improves the state of the art in each problem. Empirically, we show that the outcomes of our proposed methods can be applied in real robots and allow them to perform useful manipulation tasks.

\end{abstract}

\cleardoublepage

\addcontentsline{toc}{chapter}{Acknowledgements}

\begin{acknowledgements}

First and foremost, I sincerely thank my advisors Darwin G. Caldwell and Nikos G. Tsagarakis for continuous support to my research and for patient correction of my work. I have learnt a lot from them during these past three years, not only knowledge but also professionalism in the work. 

Besides, I would like to express my most sincere gratitude to all members of ADVR team for a great working atmosphere with lots of support they have daily shared with me. I would not be able to complete my robotic experiments without the help from Luca Muratore, Dimitrios Kanoulas, Enrico Mingo Hoffman, Alessio Rocchi, Lorenzo Baccelliere, and many other people from the team.

I also would like to thank Ian Reid and Thank-Toan Do for their support during my time at ACRV. I have learned a lot from their motivation, enthusiasm, and immense knowledge. 

A special thank to all my friends at ADVR to make every day in my last three years unforgettable:  
Brian Delhaisse, Domingo Esteban, Małgosia Kameduła, Rajesh Subburaman, Songyan Xin, Vishnu Dev, Yangwei You, Zeyu Ren.

A big thank to my parents, brother and sister who always believe and stay by my side.

\end{acknowledgements}

\body

\chapter{Introduction}

%\section{Motivation and Objectives}
\section{Motivation}

Deep learning has become a popular approach to tackle visual problems. Traditional tasks in computer vision such as object detection~\citep{Shaoqing2015} and instance segmentation~\citep{Kaiming17_MaskRCNN_short} have achieved many mini-revolutions. Despite these remarkable results, the way these problems are defined prevents them from being widely used in robotic applications. It is because the field of computer vision focuses more on the understanding step, while in robotics, the robots not only need to understand the environment but also be able to interact with it. In this work, we propose different approaches to bridge the gap between computer vision and robotics. Our long-term goal is to make the problems in computer vision become more realistic and useful for robotic applications. \\

In robotics, detecting object affordances~\citep{Gibson79} is an essential capability that allows a robot to understand and autonomously interact with objects in the environment~\citep{zech2017computational}. Most of the prior works on affordance detection have focused on grasp detection~\citep{Kragic09} using RGB-D images or point cloud data. While these methods can lead to successful grasping actions, they cannot provide other visual information for the robot to use the object as human. Unlike the visual or physical properties that mainly describe the object alone, affordances indicate functional interactions of object parts with humans. In practice, object affordances provide the key information for the manipulation tasks. For example, to pour the water from a bottle into a bowl, the robot not only has to detect the relevant objects (i.e., \textsl{bottle}, \textsl{bowl}), but also be able to localize their affordances (i.e., \texttt{grasp}, \texttt{contain}). In this study, we consider object affordances at \textit{pixel level} from an image. Therefore, the task of detecting object affordance can be considered as an extension of the well-known semantic image segmentation task in computer vision. However, detecting object affordances is a more difficult task than the classical semantic segmentation problem since object affordances represent the abstract concept when humans interact with the objects. Understanding object affordances would allow the robot to choose the right action for each manipulation task in a more autonomous way~\citep{zech2017computational}.

While object affordances give the robot a detailed understanding about the object, reasoning about human demonstrations will enable the robot to replicate human actions. In this work, we cast this problem as a visual video translation task: given a video, the goal is to translate this video to a command. Although we are inspired by the video captioning field~\citep{Donahue2014, Venugopalan2016}, there are two key differences between our approach and the traditional video captioning task: (i) we use the \textit{grammar-free} format in the captions for the convenience in robotic applications, and (ii) we aim at learning the commands through the demonstration videos that contain the \textit{fine-grained} human actions. The use of fine-grained classes forms a more challenging problem than the traditional video captioning task since the fine-grained actions usually happen within a short duration, and there is usually ambiguous information between these actions. 

To effectively learn the fine-grained actions in the task of translating videos to commands, unlike the traditional video and image captioning methods~\citep{yao2015describing, you16_image_captioning} that mainly investigate the use of \textit{visual attention} to improve the result, our architectures focus on learning and understanding the human action for the input video. Since our method provides a meaningful way to let the robots understand human demonstrations by encoding the knowledge in the video, it can be integrated with any learning from demonstration techniques to improve the manipulation capabilities of the robot. We show that, together with our affordance detection framework, the robot can autonomously perform various manipulation tasks by ``watching" the input video.

\newpage

\section{Contributions}

\textbf{Affordance Detection} In~\citep{Nguyen2016_Aff}, we propose the first deep learning framework to detect object affordances. The network has the encoder-decoder architecture to effectively encode the deep features from the input images. Our method sets a new benchmark on this task by improving the accuracy over 20\% in comparison with the state-of-the-art methods that use hand-designed geometric features. However, the drawback of this approach is it is not able to handle the input images with variable scales and also does not provide the object location. To address this, in~\citep{Nguyen2017_Aff} we introduce a sequential approach with two deep networks and a graphical model to localize both the object location and its affordances. However, this approach cannot be trained end-to-end. In~\citep{AffordanceNet17}, we propose AffordanceNet, which is an end-to-end framework that not only improves the overall performance but also enables real-time inference. Furthermore, we also demonstrate that the outputs of our affordance network can be used in various robotic manipulation applications. 

\textbf{Fine-grained Action Understanding} In~\citep{Nguyen_V2C_ICRA18}, we form the fine-grained action understanding task as a video captioning task. By using a network with two RNN layers, we can automatically translate a demonstration video to a command that can be directly used in robotic applications. Although this approach achieves the state-of-the-art translation results, it is not able to effectively encode the fine-grained action in the demonstration video. To overcome this limitation, we propose V2CNet~\citep{Nguyen_V2CNet}, a new deep learning framework that has two branches and aims at understanding the demonstration video in a fine-grained manner. The first branch of V2CNet has the encoder-decoder architecture to encode the visual features and sequentially generate the output words as a command, while the second branch uses a Temporal Convolutional Network (TCN) to explicitly learn the fine-grained actions. By jointly training both branches, the network is able to model the sequential information of the command, while effectively encodes the fine-grained actions. The experimental results on a large-scale dataset show that V2CNet outperforms recent state of the art by a substantial margin, while its output commands can be used in the real-world robotic applications. 

In both problems, we also introduce a new large-scale dataset that is suitable for deep learning. To encourage further research, we release our source code and trained models that allow reproducing the results in our papers. Finally, we also perform experiments on different robotic platforms to show the real life application of our approaches.

Apart from these two main topics, this thesis also makes contribution to other problems related to vision and manipulation:
\begin{itemize}
\item In~\citep{nguyen2016preparatory}, we propose a new method to reorient the object its nomial pose so the robot can grasp and use the object in the later tasks.
\item In~\citep{nguyen2017_event_pose}, we propose a real-time method to relocalize the 6DOF pose of the event camera using stacked spatial LSTM network. 
\item In~\citep{nguyen2018_object_captioning}, we introduce two deep frameworks that can describe the properties of the object using natural language, or retrieve the object based on an input language query.
\end{itemize}

\section{Outline}
In this dissertation, we develop neural network architectures to address the problem of affordance detection and translating videos to commands.

In Chapter \ref{ch_dl_background}, we briefly provide the fundamental background on neural network, deep learning, and how to train a deep network.

In Chapter \ref{chapter_affordance_detection}, we present our approaches to tackle the affordance detection problem. We first discuss the motivation and the related work. Then we introduce three architectures: Encoder-Decoder; Sequential; and AffordanceNet architecture for this problem. Finally, we present our experimental results using publicly available datasets as well as our robotic manipulation applications.

In Chapter \ref{ch_act_understanding}, we present two architectures (S2SNet and V2CNet) for the problem of translating videos to commands. We show that our V2CNet that uses the Temporal Convolutional Networks (TCN) to learn the fine-grained human actions is an effective way to handle this task. Lastly, we also describe our new dataset and compare our results with the state-of-the-art methods in the field, as well as the robotic applications.

Finally, in Chapter \ref{ch_conclusion} we identify the remaining challenges and discuss the future work.

\chapter{Deep Learning Background}
\label{ch_dl_background}
This chapter presents the fundamental background on artificial neural networks. We mainly focus on Convolutional Neural Networks (CNN), Recurrent Neural Networks (RNN), and how to train a deep network. For a more in-depth discussion, we prefer the readers to a recent Deep Learning Book by~\cite{goodfellow2016deep}.

\section{Neural Networks}
\subsection{Vanilla Neural Networks}
Inspired by biological neurons, artificial neurons receive input from multiple sources and output a signal based on the inputs and its activation threshold. Mathematically, an artificial neuron represents a nonlinear function $f:X \to Y$ that maps an input space $X$ to an output space $Y$.

More particular, the function $f$ can be parameterized by a \textit{weight vector} $\mathbf{w}$, a \textit{bias} $\mathbf{b}$, and a non-linear \textit{activation function} $\sigma$ as follows: 
\begin{equation}
f(\mathbf{x}) = \sigma ({\mathbf{w}^T}\mathbf{x} + \mathbf{b})
\end{equation}

Since a single neuron alone cannot represent high dimensional input data, we usually stack a set of neurons into a \textit{layer}. A multiple consecutive layers that receive the input from its previous layer and output the signal to the following layer is called a \textit{vanilla neural network}. For example, a 3 layers network could be implemented as $f(\mathbf{x}) = {\mathbf{W}_3}\sigma ({\mathbf{W}_2}\sigma ({\mathbf{W}_1}\mathbf{x}))$, where $\mathbf{W}$ is the weight matrix. The first layer of the network is called the \textit{input layer}, the last layer is the \textit{output layer}, and other layers are \textit{hidden layers}. In practice, the output layer normally does not contain the activation function as it is used to predict the groundtruth.

\subsection{Convolutional Neural Networks}
The vanilla networks only consider the input vector $\mathbf{x}$ as \textit{one} dimensional array. In \textit{Convolutional Neural Networks} (CNN), the input is instead a multi-dimensional array (i.e. a \textit{tensor}). For example, an image can be presented as 3 channels ($width \times height \times depth$) as the input for CNN. Due to the nature of its design, CNN is well suitable to handle data with spatial topology such as images, videos, and 3D voxel data.

\textbf{Convolutional Layer}
The convolutional layer is the core building block of a CNN. The main role of convolutional layer is to learn a compact representation of the high dimensional input data. This compact representation is usually called as \textit{deep feature} and widely used in many tasks. In practice, the convolutional layer compresses the input data smaller in term of width and height, but expands the data in term of depth. For example, an $300 \times 200 \times 3$ image can be transformed into a $5 \times 5 \times 1024$ deep feature vector. Each channel in this deep feature vector can be considered as a ``view" of the image, emphasizing some aspects, and de-emphasizing others.

Given a tensor as the input, the convolutional layer convolves (slides) over all spatial locations and compute dot products to output an activation map. The size of this map is controlled by three parameters \textit{depth}, \textit{stride} and \textit{zero-padding}:
\begin{itemize}
\item Depth: Depth is the number of filters we use for the convolution operation. For example, if the input of the convolutional layer is a raw image, then different neurons along the depth dimension may activate in presence of various oriented edges, or blobs of color. 

\item Stride: Stride is the number of pixels that we move the filter matrix over the input. For example, when the stride is $2$, then the filters jump 2 pixels at a time as we slide them around. Having a larger stride will produce smaller feature maps.

\item Zero-padding: In practice, it is convenient to pad the input matrix with zeros around the border, so that we can apply the filter to bordering elements of our input image matrix. The use of zero padding technique allows us to control the size of the feature maps.
\end{itemize}

\textbf{Pooling}
Pooling layer is used to reduce the dimensions of the activation map to avoid overfitting. The pooling layer operates on each 
activation map independently and downsamples them spatially. Two popular pooling techniques are \textit{max pooling} and \textit{average pooling}, both of which operate on a rectangular neighborhood. Unlike the convolutional layer, the pooling layer does not have any network parameter.

\textbf{CNN Architectures}. A CNN is built by stacking convolutional layers and (possibly) pooling layers to control the number of parameters of the network. Most of CNN architectures are designed using practical observations. Currently, the popular CNN architectures are VGG16~\citep{SimonyanZ14}, Inception~\citep{Szegedy16_Inception}, and ResNet50~\citep{He2016}.

\newpage

\subsection{Recurrent Neural Networks}
In many applications, the input or output data are sequences, e.g., a sequence of words in a sentence or a sequence of frames in a video. While CNN can effectively encode the feature in each separated frame, it cannot handle the temporal relationship in the data. Recurrent Neural Networks (RNN) are a form of neural networks designed to work with sequential data. The key idea of RNN is it keeps an internal state that is updated for each input item in the sequence. This allows RNN to retain context when processing a sequence of data. Two popular RNN are Long Short Term Memory (LSTM)~\citep{hochreiter1997long} and Gated Recurrent Unit (GRU)~\citep{Cho14_GRU}.

\textbf{LSTM} LSTM is a popular RNN since it can effectively model the long-term dependencies from the input data through its gating mechanism. The LSTM network takes an input $\mathbf{x}_t$ at each time step $t$, and computes the hidden state $\mathbf{h}_t$ and the memory cell state $\mathbf{c}_t$ as follows:

\begin{equation}
\label{Eq_LSTM} 
\begin{aligned} 
{\mathbf{i}_t} &= \sigma ({\mathbf{W}_{xi}}{\mathbf{x}_t} + {\mathbf{W}_{hi}}{\mathbf{h}_{t - 1}} + {\mathbf{b}_i})\\
{\mathbf{f}_t} &= \sigma ({\mathbf{W}_{xf}}{\mathbf{x}_t} + {\mathbf{W}_{hf}}{\mathbf{h}_{t - 1}} + {\mathbf{b}_f})\\
{\mathbf{o}_t} &= \sigma ({\mathbf{W}_{xo}}{\mathbf{x}_t} + {\mathbf{W}_{ho}}{\mathbf{h}_{t - 1}} + {\mathbf{b}_o})\\
{\mathbf{g}_t} &= \phi ({\mathbf{W}_{xg}}{\mathbf{x}_t} + {\mathbf{W}_{hg}}{h_{t - 1}} + {\mathbf{b}_g})\\
{\mathbf{c}_t} &= {\mathbf{f}_t} \odot {\mathbf{c}_{t - 1}} + {\mathbf{i}_t} \odot {\mathbf{g}_t}\\
{\mathbf{h}_t} &= {\mathbf{o}_t} \odot \phi ({\mathbf{c}_t})
\end{aligned}
\end{equation}
where $\odot$ represents element-wise multiplication, the function $\sigma: \mathbb{R} \mapsto [0,1], \sigma (x) = \frac{1}{{1 + {e^{ - x}}}}$ is the sigmod non-linearity, and $\phi: \mathbb{R} \mapsto [ - 1,1], \phi (x) = \frac{{{e^x} - {e^{ - x}}}}{{{e^x} + {e^{ - x}}}}$ is the hyperbolic tangent non-linearity. The parameters $\mathbf{W}$ and $\mathbf{b}$ are trainable weight and bias of the LSTM network. With this gating mechanism, the LSTM network can remember or forget information for long periods of time, while is still robust against the vanishing gradient problem. In practice, the LSTM network is straightforward to train end-to-end and is widely used in many problems~\citep{Donahue2014, Ramanishka2017cvpr}.

\textbf{GRU} The main advantage of the GRU network is that it uses less computational resources in comparison with the LSTM network, while achieves competitive performance. Unlike the standard LSTM cell, in a GRU cell, the update gate controls both the input and forget gates, while the reset gate is applied before the nonlinear transformation as follows:

\begin{equation}
\label{Eq_GRU} 
\begin{aligned} 
{\mathbf{r}_t} &= \sigma ({\mathbf{W}_{xr}}{\mathbf{x}_t} + {\mathbf{W}_{hr}}{\mathbf{h}_{t - 1}} + {\mathbf{b}_r})\\
{\mathbf{z}_t} &= \sigma ({\mathbf{W}_{xz}}{\mathbf{x}_t} + {\mathbf{W}_{hz}}{\mathbf{h}_{t - 1}} + {\mathbf{b}_z})\\
{\mathbf{{\tilde h}}_t} &= \phi ({\mathbf{W}_{xh}}{\mathbf{x}_t} + {\mathbf{W}_{hh}}({\mathbf{r}_t} \odot {\mathbf{h}_{t - 1}}) + {\mathbf{b}_h}\\
{\mathbf{h}_t} &= {\mathbf{z}_t} \odot {\mathbf{h}_{t - 1}} + (1 - {\mathbf{z}_t}) \odot {\mathbf{{\tilde h}}_t}
\end{aligned}
\end{equation}
where $\mathbf{r}_t$, $\mathbf{z}_t$, and $\mathbf{h}_t$ represent the reset, update, and hidden gate of the GRU cell, respectively.

\newpage

\section{Training Neural Networks}
In supervised learning for neural networks, our goal is to find a set of parameters $\theta$ of a neural network that represent a function ${f^*}()$ that best fit the input data $\mathbf{x}$ with the output data $\mathbf{y}$. During training, the network predicts $\mathbf{y^*}$ for inputs $\mathbf{x}$ from the training set, then this prediction is compared to their
corresponding groundtruth $\mathbf{y}$ to calculate the training error using the loss function $L$. We optimize this loss over all the $n$ training examples to achieve a good approximation for function ${f^*}()$:
\begin{equation}
{f^*}(\theta) \approx \mathop {argmin}\limits_{f \in F} \frac{1}{n}\sum\limits_{i = 1}^n {L(f(\mathbf{x_i}),\mathbf{y_i})} + R(f)
\end{equation}

\textbf{Backpropagation}: To minimize the loss $L$ and update the network parameters $\theta$, a popular technique is backpropagation~\citep{rumelhart1986learning}. The backpropagation algorithm is a recursive application of the chain rule that we efficiently compute gradients of scalar valued functions with respect to their inputs.

In practice, it is usually not feasible to compute the training error from the predictions for all training examples due to memory limitation, therefore, a random subset of all data (i.e. \textit{minibatch}) is used to compute the updates to the network parameters~\citep{rumelhart1986learning}.

One of the critical problem when training a neural network is to avoid overfitting, i.e., the network performs poorly on previously unseen data during testing. Regularization is an essential technique that should be used when training a neural network to combat overfitting. The popular regularization methods are Weight Decay, Dropout~\citep{srivastava2014dropout}, and Batch Normalization~\citep{ioffe2015batch}:

\begin{itemize}
\item \textbf{Weight Decay} Weight decay or $L^2$ regularization is a regularization technique in machine learning that can be used for neural network. In practice, the weight decay coefficient is added in the loss function to determine how dominant this regularization term will be in the gradient computation. When the weight decay coefficient is big, the penalty for big weights of the network is also big, when it is small weights can freely grow.

\item \textbf{Dropout} is a popular technique to combat overfitting during training by randomly disabling a neuron and its connections. This can be done by setting the weight of these neurons to zero. Intuitively, disabling neurons prevents layers from relying on specific inputs too much, thus requiring them to better generalize by utilizing more of its inputs. In practice, dropout is cheap to compute and more effective than weight decay regularization.

\item \textbf{Batch Normalization} When training a neural network, batch normalization allows each layer of a network to learn more independently of other layers. To achieve this, batch normalization technique normalizes the output of a previous activation layer by subtracting the batch mean and dividing by the batch standard deviation. Then it adds two trainable parameters to each layer, so the normalized output is multiplied by a standard deviation parameter and add a mean parameter. After this step, the scale of input features from layers would not extremely different, hence avoiding numerical issues.

\end{itemize}

\chapter{Affordance Detection}
\label{chapter_affordance_detection}

\section{Introduction}

An object can be described by various visual properties such as color, shape, or physical attributes such as weight, volume, and material. Those properties are useful to recognize objects or classify them into different categories, however they do not imply the potential actions that human can perform on the object. The capability to understand functional aspects of objects or \textit{object affordances} has been studied for a long time~\citep{Gibson79}. Unlike other visual or physical properties that mainly describe the object alone, affordances indicate functional interactions of object parts with humans. Understanding object affordances is, therefore, crucial to let an autonomous robot interact with the objects and assist humans in various daily tasks.

\begin{figure}[!t] 
\vspace{0.2cm}
    \centering 
 	\includegraphics[width=0.99\linewidth, height=0.19\linewidth]{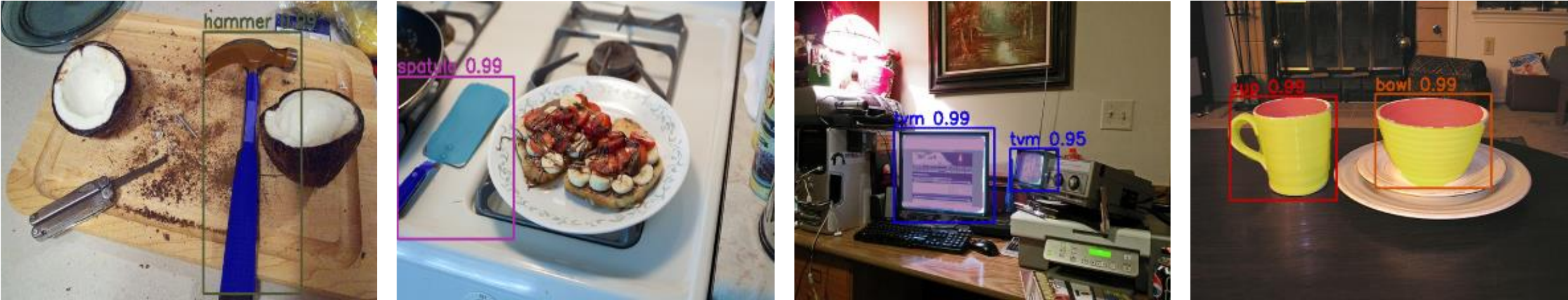} 
    \vspace{0.0ex}
    \caption{Simultaneous affordance and object detection.}
    \label{Fig_aff_intro} 
    
\end{figure}

The problem of modeling object affordances can be considered in different ways. ~\cite{Castellini2011_short} defined affordances in terms of human hand poses during the interaction with objects, while in~\citep{Koppula:2016_short} the authors studied object affordances in the context of human activities. In this dissertation, similar to~\citep{Myers15}, we consider object affordances at \textit{pixel} level from an  image, i.e., a group of pixels which shares the same object functionality is considered as one affordance. The advantage of this approach is we can reuse the strong state of the art from the semantic segmentation field, while there is no extra information such as interactions with human is needed. Detecting object affordances, however, is a more difficult task than the classical semantic segmentation problem. For example, two object parts with different appearances may have the same affordance label. Furthermore, in order to be used in the real robot, it is also essential for an affordance detection method to run in real-time and generalize well on unseen objects. Fig.~\ref{Fig_aff_intro} shows some example results of one of our proposed approach (i.e., AffordanceNet).

In particular, from the input 2D image, our goal is to detect the objects (including the object positions and its class) and their affordances. Follow the standard design in computer vision, the object position is defined by a rectangle with respect to the top-left corner of the image; the object class is defined over the rectangle; the affordances are encoded at every pixel inside the rectangle. The region of pixels on the object that has the same functionality is considered as one affordance. Ideally, we want to detect all relevant objects in the image and map each pixel in these objects to its most probable affordance label.

\section{Related Work}

The concept of object affordance has been extensively studied in robotics and computer vision over the last few years. We refer the readers to~\citep{zech2017computational} for a detailed discussion about affordance from a robotic point of view, and~\citep{hassanin2018visual} for a recent survey about visual affordance. Here, we summarize two most related fields to our work: affordance for grasping and object affordance at pixel-level.

\textbf{Grasp Affordance} Affordance related to grasping is a well-known problem in robotics~\citep{Kragic09}. In~\citep{Montesano09} the authors proposed a method to detect grasp affordances by learning a mapping from local visual descriptors to grasp parameters. In~\citep{Aldoma12} a set of the so-called $0$-ordered affordances is detected from the full 3D object mesh models. The work in~\citep{Yuyin13} proposed a method to identify color, shape, material, and name attributes of objects selected in a bounding box from RGB-D data, while in~\citep{Moldovan14} the authors introduced the concept of relational affordances to search for objects in occluded environments.  In~\citep{Schoeler2016_short}, the authors used predefined primary tools to infer object functionalities from 3D point clouds. The work in~\citep{Song15} proposed to combine the global object poses with its local appearances to detect grasp affordances. In \citep{Hedvig2011}, the authors introduced a method to detect object affordances via object-action interactions from human demonstrations. Recently,  many methods used deep learning methods to detect grasp location. For example, the work in~\citep{Lenz14} proposed a method to detect grasp affordance using two deep networks. Similarly, Levine et al.~\citep{Sergey16} collected a large amount of grasping data and applied a deep network to learn successful grasps from monocular images.

Besides the traditional grasping problem, many works have investigated different problems such as using tools from detected affordances~\citep{Dehban2016}, exploring actions and effects when robot interacts with objects~\citep{Ugur2015}, or reorienting objects for task-oriented grasping~\citep{Nguyen2016_Re}.

%Many works have focused on localizing grasp location on objects using vision~\citep{Kragic09}~\citep{Andreas15}. The work in~\citep{Lenz14_short} used two deep neural networks to detect grasp affordances from RGB images.  
%The authors in~\citep{Mar2015} proposed a method to learn tool affordances by clustering the effects of robot's actions and applied it to a humanoid platform. 

\textbf{Pixel-Level Affordance} The problem of understanding affordances at the pixel level has been termed ``object part labelling'' in the computer vision community, while it is more commonly known as ``affordance detection'' in robotics. In computer vision, the concept of affordances is not restricted to objects, but covers a wide range of applications, from understanding human body parts~\citep{Lin:2017:RefineNet} to environment affordances~\citep{Roy2016_short}, while in robotics, researchers focus more on the real-world objects that the robot can interact with. For example, in~\citep{Myers15}, the authors used hand-designed geometric features to detect object affordances at pixel level from RGB-D images.

With the rise of deep learning, recent works relied on deep neural networks for designing affordance detection frameworks.  The work in~\citep{Roy2016_short} introduced multi-scale CNN to localize environment affordances. In~\citep{Sawatzky2017}, to avoid depending on costly pixel groundtruth labels, a weakly supervised deep learning approach was presented to segment object affordances. In computer vision, Badrinarayanan et al.~\citep{Badrinarayanan15} used a deep CNN with an encoder-decoder architecture for real-time semantic pixel-wise labeling. The work in~\citep{Chen2016_deeplab} combined CNN with dense CRF to further improve the segmentation results. The work of~\citep{Li2016_FCIS} introduced an end-to-end architecture to simultaneously detect and segment object instances. Recently, the authors in~\citep{Kaiming17_MaskRCNN_short} improved over~\citep{Li2016_FCIS} by proposing a region alignment layer which effectively aligns the spatial coordinates of region of interests between the input image space and the feature map space. 

%In computer vision, following the work in~\citep{Alex12} on image classification, CNN has become very popular and been applied in many other problems. Ren et al.~\citep{Shaoqing2015} proposed a method to combine a deep CNN with region proposal methods for detecting objects. Similarly, the work in~\citep{Dai2016} developed a method that integrated region proposals with the very deep network ResNets~\citep{He2016} for object detection. 

\newpage

\section{Encoder-Decoder Architecture}\label{sec_encoder_decoder}

\subsection{Data Representation}
Recently, many works in computer vision and machine learning have investigated the effectiveness of using multiple modalities as inputs to a deep network, such as video and audio~\citep{Ngiam11} or RGB-D data~\citep{Gupta14}. However, the problem of picking the best combination of these modalities for a new task is still an open problem. Ideally, they should represent important properties of the data so that the network can effectively learn deep features from them.

\begin{figure}[ht] 
    \centering
	\includegraphics[width=0.9\linewidth, height=0.55\linewidth]{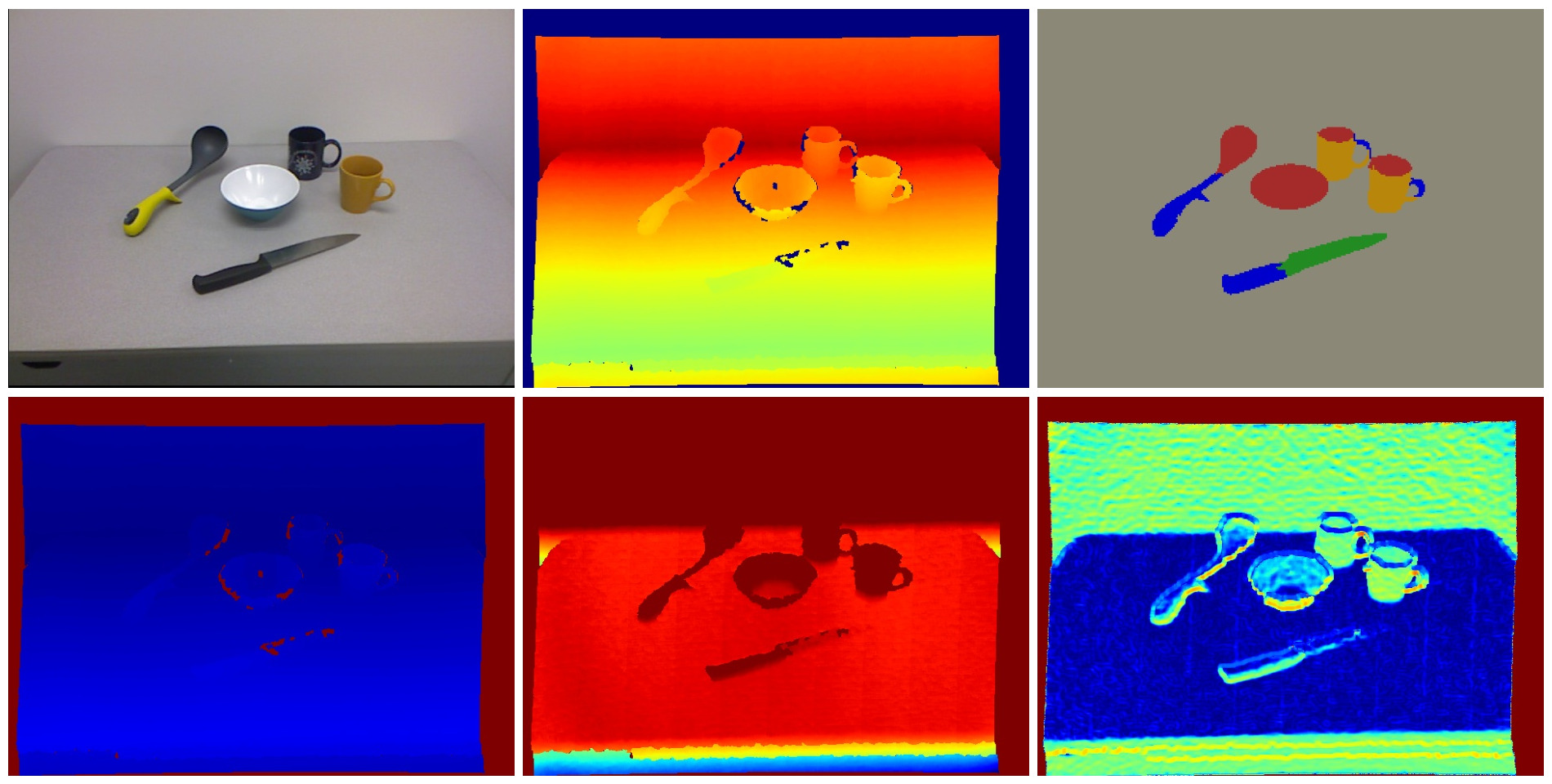}
    \vspace{0ex}
    \caption{Data representation. \textbf{Top row:} The original RGB image, its depth image, and the ground-truth affordances, respectively. \textbf{Bottom row:} The HHA representation of a depth image.}
    \label{Fig:data_rep} 
\end{figure}

\begin{figure*}[ht] 
    \centering

	\includegraphics[scale=0.73]{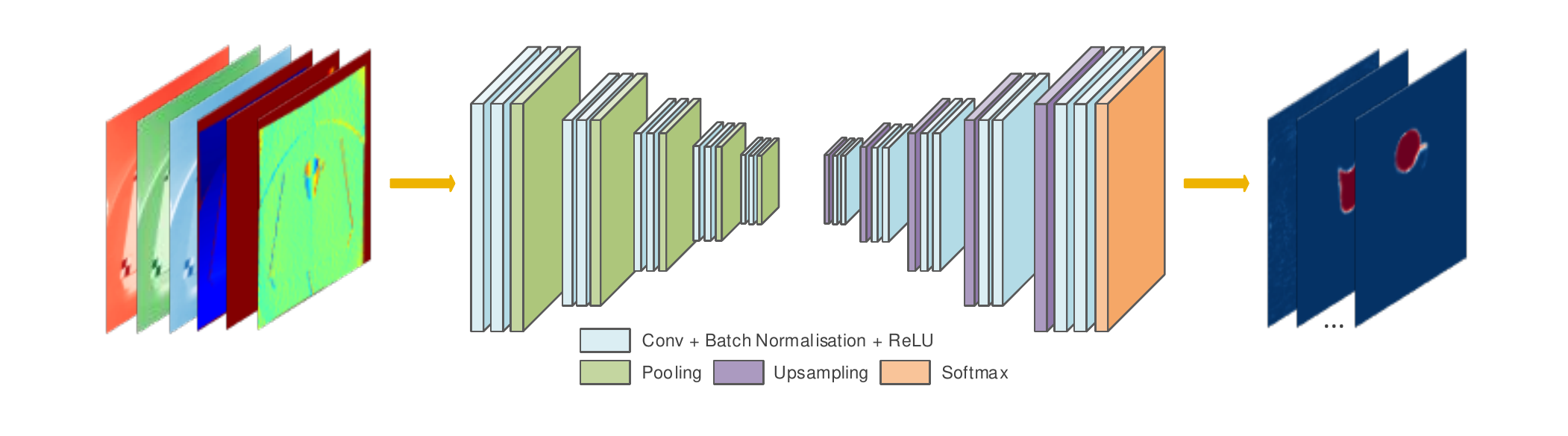}    				%\includegraphics[width=0.99\linewidth, height=0.2\linewidth]{example-image-a} 
    \vspace{-0.5ex}
    \caption{An illustration of our Encoder-Decoder Architecture. \textbf{From left to right:} The input data are represented as multiple modalities and learned by a CNN with an encoder-decoder architecture. The CNN produces a $k$ channel image of probabilities, where \textit{k} is the number of affordance classes. Each channel is visualized as an image in this figure.}
    \label{Fig:exp_architecture} 
\end{figure*}

Intuitively, we can either use only RGB images or combine both RGB and their associated depth images as the input to our network. In this work we also investigate other ways of data representation that may improve further the performance. In~\citep{Gupta14}, the authors showed that when the training data is limited (which is true in our case since the affordance dataset~\citep{Myers15} that we use for training has only 30,000 images, compared to other ones that are deep learning oriented with million of images~\citep{Alex12}), it is unlikely that the CNN would automatically learn important depth properties. To deal with this problem a new method ~\citep{Gupta14} was proposed to encode the depth images into three channels at each pixel: the horizontal disparity, the height above the ground, and the angle between each pixel's surface normal and direction of inferred gravity (denote as HHA). The HHA encoder is calculated based on an assumption that the direction of gravity would impose important information about the environment structure. We adapted this representation since the experimental results in~\citep{Gupta14} have shown that the features can be learned more effectively for object recognition tasks in indoor scenes. We show an example of different data representations for our network in Fig.~\ref{Fig:data_rep}.

\subsection{Architecture}
In 2012, the authors of~\citep{Alex12} used CNN for classifying RGB images and showed substantially higher accuracy over the state-of-the-art. Many works have applied CNN to different vision problems since then~\citep{SimonyanZ14, KaimingHe15}. Nonetheless, the design of a CNN for image segmentation still remains challenging. More recently, the work of~\citep{Noh15} proposed an encoder-decoder architecture for pixel-wise image labeling. However, the encoder of this work includes the fully connected layers that make the training very difficult due to a huge amount of parameters (approximately 134M), and also significantly increases the inference time. The authors in~\citep{Badrinarayanan15}  pursued the same idea but they discarded the fully connected layers to reduce the number of parameters. They showed that the encoder-decoder architecture without fully connected layers can still be trained end-to-end effectively without sacrificing the performance and enabling real-time reference. 

In this work, we use the state-of-the-art deep convolutional network described in~\citep{Badrinarayanan15}. In particular, the network contains two basic components: the encoder and the decoder network. The encoder network has $13$ convolutional layers that were originally designed in the VGG16 network~\citep{SimonyanZ14} for object classification. Each encoder has one or more convolutional layers that perform batch normalization, ReLU non-linearity, followed by a non-overlapping max-pooling with a $2\times 2$ window to produce a dense feature map.  Each decoder layer is associated with an encoder one, ending up in a $13$ layers decoder network. In each one, the input feature map is upsampled using the memorized pooled indices and convoled with a trainable filter bank.  The final decoder layer produces the high dimensional features that are fed to a multi-class soft-max layer, which classifies each pixel independently.  The output of the softmax layer is a \textit{k} channel image of probabilities, where \textit{k} is the number of classes. 

We adapt the above architecture to detect object affordances at pixel level. Fig.~\ref{Fig:exp_architecture} shows an overview of our approach. The data layer is modified to handle multiple modalities as input, while each image in the training set is center cropped on all channels to $240\times320$ size from its original $480\times640$ size. In testing step, we don't crop the images but use the sliding window technique to move the detected window over the test images. The final predicted result corresponds to the class with the maximum probability at each pixel over all the sliding windows. Finally, since the dataset that we use has a large variation in the number of pixels for each class in the training set, we weigh the loss differently based on this number.

\subsection{Training}
For the training, we generally follow the procedure described in~\citep{Badrinarayanan15} using the Caffe library~\citep{Jia14}. Given that the gradient instability in the deep network can stall the learning, the initialization of the network weights is very important. In particular, we initialized the network using the technique described in~\citep{KaimingHe15}.  The network is end-to-end trained using stochastic gradient descent with a $0.9$ momentum.  The cross-entropy loss~\citep{Long14} is used as the objective function for the network. The batch size was set to $10$ while the learning rate was initialized to $0.001$, and decreased by a factor of $10$ every $50,000$ iterations. The network is trained from scratch until convergence with no further reduction in training loss. The training time is approximately $3$ days on an NVIDIA Titan X GPU.

\section{Sequential Architecture}\label{sec_sequential}

\begin{figure*}[!t] 
    \centering

	\includegraphics[scale=0.63]{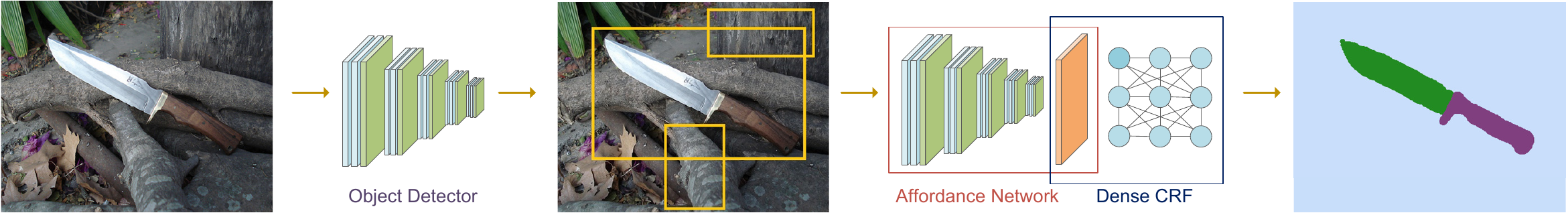}    				
    \vspace{1.5ex}
    \caption{An overview of our Sequential Architecture. \textbf{From left to right:} A deep network is first used as an object detector to generate the object bounding boxes that narrow down the region of interest. A second network is then used to produce feature maps from these bounding boxes. Finally, these maps are post-processed with dense CRF to improve the prediction along the class boundaries.}
    \label{fig_seq_overview} 
\end{figure*}

Although the Encoder-Decoder Architecture (Section ~\ref{sec_encoder_decoder}) effectively outputs the affordance map as the result, it obmits the object location. This would become a major drawback in many robotic applications when the object location is needed. Therefore, in this Sequential Architecture (Fig.~\ref{fig_seq_overview}), we address the problem of integrating object detector to an affordance detection system. 

\subsection{Object Detector} Detecting objects in RGB images is a well-known problem in computer vision. Recently, rapid advancements have been made in this field with the rise of deep learning. Object detection approaches can be divided into region-based~\citep{Shaoqing2015}~\citep{Dai2016} and non region-based methods~\citep{Wei2016}. Although non region-based methods can achieve real-time performance, they are still outperformed by region-based systems on public benchmarks~\citep{Dai2016}.

We first use the method that was proposed recently in~\citep{Dai2016} (R-FCN) as our object detector since it achieves state-of-the-art results on public benchmarks, and has fast inference time. Unlike other object detection methods that modified the VGG-16~\citep{SimonyanZ14} network in their main architecture, R-FCN is designed to naturally adopt the state-of-the-art image classification network, such as ResNets~\citep{He2016}, hence letting us train a very deep network in a fully convolutional manner. R-FCN contains two main steps: generating region proposals and classifying them. The candidate regions are first generated by a deep network that shares features with R-FCN, then the training process classifies each region into object categories and background. During testing, a threshold $t$ is chosen to decide whenever a region proposal belongs to an object category or not. We refer readers to~\citep{Dai2016} for the full description of the R-FCN architecture.

\subsection{Architecture}
Similar to~\citep{Abhilash2016, Nguyen2016_Aff, Roy2016}, we cast the affordance detection problem as a pixel-wise labeling task. We modify the state-of-the-art VGG-16 network to produce dense feature map at each object region provided by R-FCN. Since the original VGG-16 network is designed for the image classification problem, its final layer is a classifier and can't produce dense heat maps to predict the affordance label for each pixel. Therefore, we replace this layer with an $1 \times 1$ convolution layer of $10$ dimensions to predict scores for each class in our dataset (we use a dataset with $9$ affordance classes and $1$ class for the background). We then convert all the fully-connected layers of VGG-16 into convolutional ones. Furthermore, we use atrous convolution technique~\citep{Chen2016_deeplab} to increase the field-of-view of the convolution layers without increasing the number of network parameters. This technique also helps us to balance the trade-off between small field-of-view for accurate localization and large field-of-view for incorporating context information.

To deal with arbitrary resolutions from the input images, we apply a multi-scale strategy introduced in~\citep{Iasonas2015}. During the training and testing, we re-scale the original image into three different versions and feed them to three parallel networks that share the same parameters. The final feature map is created by bilinearly interpolating the feature map from each network to the original image resolution, and taking the maximum value across the three scales at each pixel.

\subsection{Post-processing with CRF}
We adopt dense CRF to post-process the output from the deep network since it showed substantial performance gains in traditional the image segmentation task~\citep{Chen2016_deeplab, Zheng2015}. The energy function of dense CRF is given by:

\begin{equation}
E(\mathbf{x}|\mathbf{P}) = \sum\limits_\mathbf{\mathbf{p}} {\theta _\mathbf{p}({x_\mathbf{p}}) + \sum\limits_{\mathbf{p},\mathbf{q} } {\psi _{\mathbf{\mathbf{p}},\mathbf{q}}({x_\mathbf{p}},{x_\mathbf{q}})} }
\end{equation}

In particular, the unary term $\theta _\mathbf{p}({x_\mathbf{p}})$ indicates the cost of assigning label $x_\mathbf{p}$ to pixel $\mathbf{p}$. This term can be considered as the output of the last layer from the affordance network since this layer produces a probability map for each affordance class~\citep{Chen2016_deeplab}. The pairwise term $\psi _{\mathbf{\mathbf{p}},\mathbf{q}}({x_\mathbf{p}},{x_\mathbf{q}})$ models the relationship among neighborhood pixels and penalizes inconsistent labeling. The pairwise potential can be defined as weighted Gaussians:

\begin{equation}
\psi _{\mathbf{p},\mathbf{q}}({x_\mathbf{p}},{x_\mathbf{q}}) = \mu ({x_\mathbf{p}},{x_\mathbf{q}})\sum\limits_{m = 1}^M {{w^m}{\kappa ^m}({\mathbf{f}_\mathbf{p}},{\mathbf{f}_\mathbf{q}})} 
\end{equation}
where each $\kappa^m$ for $m=1,...,M$, is a Gaussian kernel based on the features $\mathbf{f}$ of the associated pixels, and has the weights $w^m$. The term $\mu ({x_\mathbf{p}},{x_\mathbf{q}})$ represents label compatibility and is $1$ if $x_\mathbf{p} \ne x_\mathbf{q}$, otherwise $0$. As in~\citep{Philipp2011}, we use the following kernel in the pairwise potential: 

\begin{equation}
\begin{array}{c}
\kappa ({\mathbf{f}_\mathbf{p}},{\mathbf{f}_\mathbf{q}}) = {w_1}\exp \left( { - \frac{{|{p_\mathbf{p}} - {p_\mathbf{q}}{|^2}}}{{2\sigma _\alpha ^2}} - \frac{{|{I_\mathbf{p}} - {I_\mathbf{q}}{|^2}}}{{2\sigma _\beta ^2}}} \right)\\
 + {w_2}\exp \left( { - \frac{{|{p_\mathbf{p}} - {p_\mathbf{q}}{|^2}}}{{2\sigma _\gamma ^2}}} \right)
\end{array}
\end{equation}
where the first term depends both on pixel positions (denoted as $p$) and its color (denoted as $I$), and the second term only depends on pixel positions. The parameter $\sigma$ controls the scale of the Gaussian kernel.

Our goal is to minimize the CRF energy $E(\mathbf{x}|\mathbf{P})$, which yields the most probable label for each pixel. Since the dense CRF has billion edges and the exact minimization is intractable, we use the mean-field algorithm~\citep{Philipp2011} to efficiently approximate the energy function.

\subsection{Training}

We generally follow the procedure described in~\citep{Dai2016} to train our object detector. The network is trained using gradient descent with $0.9$ momentum, $0.0005$ weight decay. The input images are resized to $600 \times 600$ pixels resolution. The learning rate is first set to $0.001$, then we decrease it by a factor of $10$ every $20000$ iterations. During training, we use $128$ regions of interest as bounding box candidates for backpropagation. The network is trained using the lost function combined from cross-entropy loss and  box regression loss. The training time is approximately $1$ day on a NVIDIA Titan X GPU.

After training the object detector network, we do the inference to generate object bounding boxes in order to feed these boxes to the affordance network. The affordance network is trained using stochastic gradient descent with cross-entropy loss, $0.9$ momentum and $0.0005$ weight decay. The learning rate is initialized to $0.001$ and decreased by a factor of $10$ every $2000$ iterations. Similar to~\citep{Nguyen2016_Aff}, we weigh the loss differently based on the statistics of each class to deal with the large variation in the number of pixels in the training set. The network is trained until convergence with no further reduction in training loss. It takes approximately $1$ day to train our affordance network on a Titan X GPU.

\section{AffordanceNet Architecture}

Although the Sequential Architecture (Section.~\ref{sec_sequential}) uses a deep learning-based object detector to improve the affordance detection accuracy on a real-world dataset. A limitation of this approach is that its architecture is not end-to-end -- i.e. two separate networks are used, one for object detection and one for affordance detection -- and this is slow for both training and testing. Furthermore, by training two networks separately, the networks are not jointly optimal. In computer vision, the work of~\citep{Li2016_FCIS} introduced an end-to-end architecture to simultaneously detect and segment object instances. Recently, the authors in~\citep{Kaiming17_MaskRCNN_short} improved over~\citep{Li2016_FCIS} by proposing a region alignment layer which effectively aligns the spatial coordinates of region of interests between the input image space and the feature map space.

The goal of this architecture is to simultaneously detect the objects (including the object location and object label) and their associated affordances. We follow the same concept in~\citep{Nguyen2017_Aff}, however we use an end-to-end architecture instead of a sequential one. Our object affordance detection network can also be seen as a generalization of the recent state-of-the-art instance segmentation networks~\citep{Kaiming17_MaskRCNN_short, Li2016_FCIS}. In particular, our network can detect multiple affordance classes in the object, instead of binary class as in instance segmentation networks~\citep{Kaiming17_MaskRCNN_short, Li2016_FCIS}.

We first describe three main components of our AffordanceNet: the Region of Interest (RoI) alignment layer (RoIAlign)~\citep{Kaiming17_MaskRCNN_short} which is used to correctly compute the feature for an RoI from the image feature map; a sequence of convolution-deconvolution layers to upsample the RoI feature map to high resolution in order to obtain a smooth and fine affordance map; a robust strategy for resizing the training mask to supervise the affordance detection branch. We show that these components are the key factors to achieve high affordance detection accuracy. Finally, we present the whole AffordanceNet architecture in details.

\subsection{RoIAlign}
One of the main components in the recent successful region-based object detectors such as Faster R-CNN~\citep{Shaoqing2015} is the Region Proposal Network (RPN). This network shares weights with the main convolutional backbone and outputs bounding boxes (RoI / object proposal) at various sizes. For each RoI, a fixed-size small feature map (e.g.,  $7\times7$) is pooled from the image feature map using the RoIPool layer~\citep{Shaoqing2015}. The RoiPool layer works by dividing the RoI into a regular grid and then max-pooling the feature map values in each grid cell. This quantization, however, causes misalignments between the RoI and the extracted features due to the harsh rounding operations when mapping the RoI coordinates from the input image space to the image feature map space and when dividing the RoI into grid cells.

In order to address this problem, the authors in~\citep{Kaiming17_MaskRCNN_short} introduced the RoIAlign layer which properly aligns the extracted features with the RoI. Instead of using the rounding operation, the RoIAlign layer uses bilinear interpolation to compute the interpolated values of the input features at four regularly sampled locations in each RoI bin, and aggregates the result using max operation. This alignment technique plays an important role in tasks based on pixel level such as image segmentation. We refer the readers to~\citep{Kaiming17_MaskRCNN_short} for a detailed analysis of the RoIAlign layer.

\subsection{Mask Deconvolution}
In recent state-of-the-art instance segmentation methods such Mask-RCNN~\citep{Kaiming17_MaskRCNN_short} and FCIS~\citep{Li2016_FCIS}, the authors used a small fixed size mask (e.g. $14 \times 14$ or $28 \times 28$) to represent the object segmentation mask. This is feasible since the pixel value in each predicted mask of RoI is binary, i.e., either foreground or background. We empirically found that using small mask size does not work well in the affordance detection problem since we have multiple affordance classes in each object. Hence, we propose to use a sequence of deconvolutional layers for achieving a high resolution affordance mask.

Formally, given an input feature map with size $S_i$, the deconvolutional layer performs the opposite operation of the convolutional layer to create a bigger output map with size $S_o$, in which $S_i$ and $S_o$ are related by:
\begin{equation}
%{S_o} = \delta*({S_i} - 1) + {S_f} - 2*\rho
{S_o} = s*({S_i} - 1) + {S_f} - 2*d
\end{equation}
where $S_f$ is the filter size; $s$ and $d$ are stride and padding parameters, respectively.
\begin{figure}[!ht] 
    \centering
	\includegraphics[width=0.95\linewidth, height=0.3\linewidth]{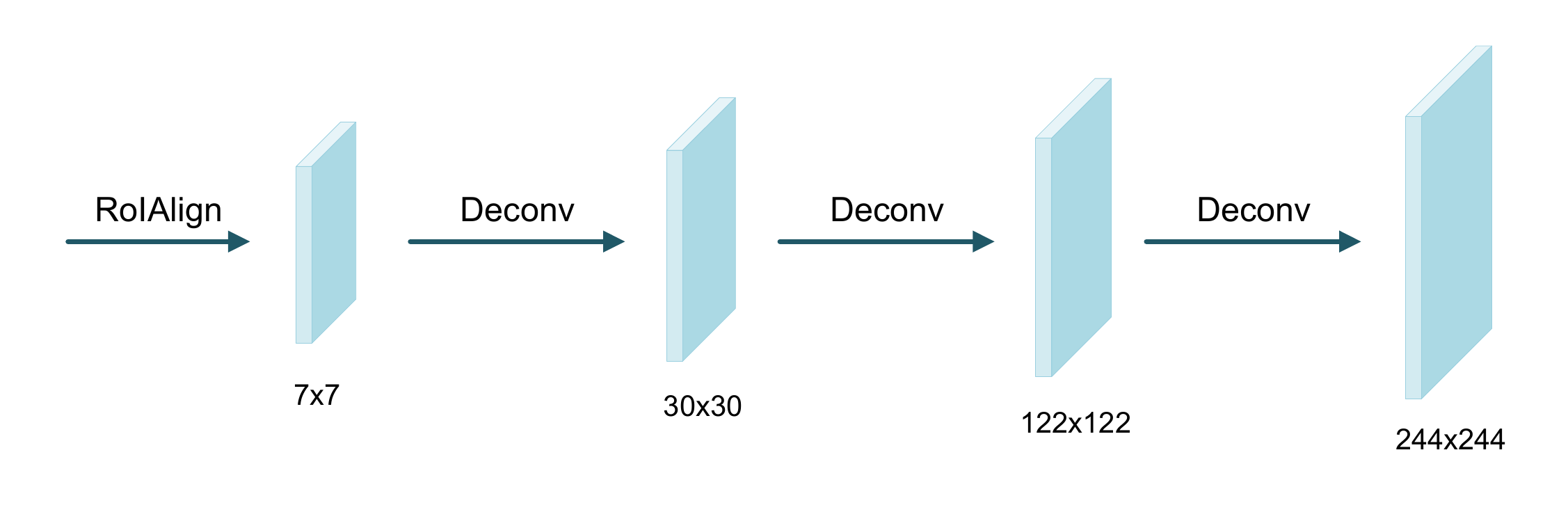}    				
    \vspace{1.5ex}
    \caption{A sequence of three deconvolutonal layers to gradually upsample a $7 \times 7$ fixed size feature map to $244 \times 244$.}
    \label{fig:three_deconvo} 
\end{figure}

In practice, the RoIAlign layer outputs a feature map with size $7 \times 7$. We use three deconvolutional layers to upsample this map to higher resolution (see Fig.~\ref{fig:three_deconvo}). The first deconvolutional layer has the padding $d=1$, stride $s=4$, and kernel size $S_f=8$ to create the map with size $30 \times 30$. Similarly, the second layer has the parameters ($d=1$, $s=4$, $S_f=8$), and the third one has ($d=1$, $s=2$, $S_f=4$) to create the final high resolution map with the size of $244 \times 244$. It is worth noting that before each deconvolutional layer, a convolutional layer (together with ReLu) is used to learn features which will be used for the deconvolution. This convolutional layer can be seen as an adaptation between two consecutive deconvolutional layers.

\subsection{Resizing Affordance Mask}
\label{subsub:resizing}
Similar to Mask-RCNN~\citep{Kaiming17_MaskRCNN_short} and FCIS~\citep{Li2016_FCIS}, our affordance detection branch requires a fixed size (e.g., $244\times 244$) target affordance mask to supervise the training. During training, the authors in~\citep{Kaiming17_MaskRCNN_short, Li2016_FCIS} resized the original groundtruth mask of each RoI to the pre-defined mask size to compute the loss. This resizing step outputs a mask with values ranging from 0 to 1, which is thresholded (e.g., at $0.4$) to determine if a pixel is background or foreground. However, using single  threshold value does not work in our affordance detection problem since we have multiple affordance classes in each object. To address this problem, we propose a resizing strategy with multi-thresholding. 

Given an original groundtruth mask, without loss of generality, let $P=\{c_0,c_1,...,c_{n-1}\}$ be set of $n$ unique labels in that mask, we first linearly map the values in $P$ to $\hat{P}=\{0, 1, ..., n-1\}$ and convert the original mask to a new mask using the mapping from $P$ to $\hat{P}$. We then resize the converted mask to the pre-defined mask size and use the thresholding on the resized mask as follows: %(e.g., $244 \times 244$). 

\begin{equation}\label{Eq_resize_map}
   \rho(x, y)=
    \begin{cases}
      \hat{p}, & \text{if}\ \hat{p} - \alpha  \le \rho(x, y) \le \hat{p} + \alpha \\
      0, & \text{otherwise}
    \end{cases}
\end{equation}
where $\rho(x, y)$ is a pixel value in the resized mask; $\hat{p}$ is one of values in $\hat{P}$; $\alpha$ is the hyperparameter and is set to $0.005$ in our experiments.

Finally, we re-map the values in the thresholded mask back to the original label values (by using the mapping from $\hat{P}$ to $P$) to achieve the target training mask.

\begin{figure*}[!t] 
	\vspace{-0.3cm}
    \centering   			
		\includegraphics[scale=0.28]{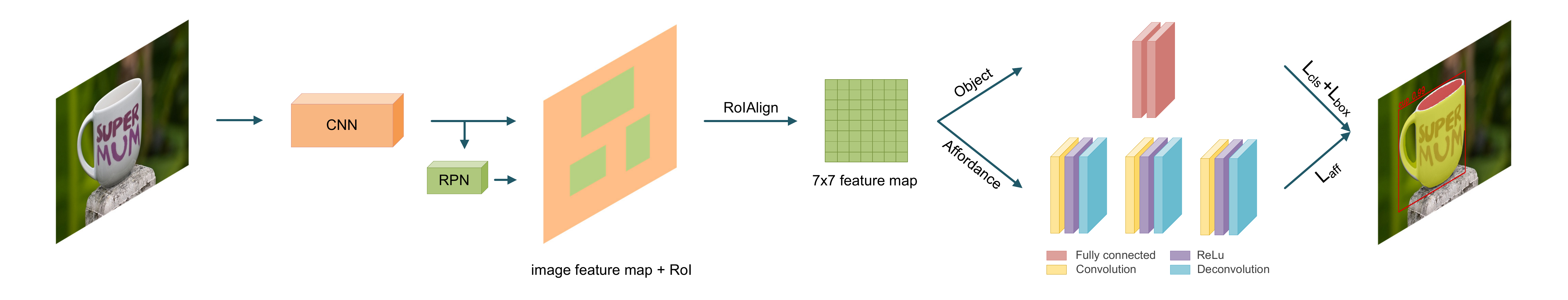}    				
    \vspace{1ex}
    \caption{An overview of our AffordanceNet. \textbf{From left to right:} A deep CNN backbone (i.e., VGG) is used to extract image features. The RPN shares weights with the backbone and outputs RoIs. For each RoI, the RoIAlign layer extracts and pools its features (from the image feature map, i.e., the $conv5\_3$ layer of VGG) to a fixed size $7\times 7$ feature map. The object detection branch uses two fully connected layers for regressing object location and classifying object category. The object affordance detection branch consists of a sequence of convolutional-deconvolutional layers and ends  with a softmax layer to output a multiclass affordance mask.}
    \label{fig:aff_overview} 
\end{figure*}

\subsection{Architecture}
Fig.~\ref{fig:aff_overview} shows an overview of our end-to-end affordance detection network. The network is composed of two branches for object detection and affordance detection. 

Given an input image, we use the VGG16~\citep{SimonyanZ14} network as the backbone to extract deep features from the image. A RPN that shares the weights with the convolutional backbone is then used to generate candidate bounding boxes (RoIs). For each RoI, the RoIAlign layer extracts and pools its corresponding features (from the image feature map --- the $conv5\_3$ layer of VGG16) into a $7 \times 7$ feature map. In the object detection branch, we use two fully connected layers, each with $4096$ neurons, followed by a classification layer to classify the object and a regression layer to regress the object location. In the affordance detection branch, the $7 \times 7$ feature map is gradually upsampled to $244 \times 244$ to achieve high resolution map. The affordance branch uses a softmax layer to assign each pixel in the $244 \times 244$ map to its most probable affordance class. The whole network is trained end-to-end using a multi-task loss function.

\subsection{Multi-Task Loss}
In our aforementioned end-to-end architecture, the classification layer outputs a probability distribution $p = (p_0,...,p_K)$ over $K+1$ object categories, including the background. As in~\citep{Shaoqing2015}, $p$ is the output of a softmax layer. The regression layer outputs $K+1$ bounding box regression offsets (each offset includes box center and box size): $t^k = (t^k_x,t^k_y,t^k_w,t^k_h)$. Each offset $t^k$ corresponds to each class $k$. Similar to~\citep{Girshick2014, Shaoqing2015} we parameterize for $t^k$, in which $t^k$ specifies a scale-invariant translation and log-space height/width shift relative to an anchor box of the RPN. The affordance detection branch outputs a set of probability distributions $m = \{m^i\}_{i \in RoI}$ for each pixel $i$ inside the RoI, in which $m^i = (m^i_0,...,m^i_C)$ is the output of a softmax layer defined on $C+1$ affordance labels, including the background. 

We use a multi-task loss $L$ to jointly train the bounding box class, the bounding box position, and the affordance map as follows:
\begin{equation}\label{loss_coarse}
L = {L_{cls}} + {L_{loc}} + {L_{aff}}
\end{equation}
where ${L_{cls}}$ is defined on the output of the classification layer; ${L_{loc}}$ is defined on the output of the regression layer; ${L_{aff}}$ is defined on the output of the  affordance detection branch.

The prediction target for each RoI is a groundtruth object class $u$, a groundtruth bounding box offset $v$, and a target affordance mask $s$. The values of $u$ and $v$ are provided with the training datasets. The target affordance mask $s$ is the intersection between the RoI and its associated groundtruth mask. For pixels inside the RoI which do not belong to the intersection, we label them as background. Note that the target mask is then resized to a fixed size (i.e., $244\times 244$) using the proposed resizing strategy in Section \ref{subsub:resizing}. 
Specifically, we can rewrite Equation~\ref{loss_coarse} as follows:
\begin{align}\label{eq:loss_fine}
L(p,u,t^u,v,m,s) =& L_{cls}(p,u) + I[u\ge1]L_{loc}(t^u,v) \nonumber \\ 
&+ I[u\ge1]L_{aff}(m,s)
\end{align}
%where the $L_{cls}(p,u) = -log(p_u)$ is the multinomial cross entropy loss for the classification; $p_u$ is the softmax output for the true class $u$. 
The first loss $L_{cls}(p,u)$ is the multinomial cross entropy loss for the classification and is computed as follows:
\begin{equation}
 L_{cls}(p,u) = -log(p_u)
\end{equation} 
where $p_u$ is the softmax output for the true class $u$. 

The second loss $L_{loc}(t^u,v)$ is \textit{Smooth L1} loss~\citep{Girshick2014} between the regressed box offset $t^u$ (corresponding to the groundtruth object class $u$) and the groundtruth box offset $v$, and is computed as follows:
\begin{equation}
L_{loc}(t^u,v) = \sum_{i\in\{x,y,w,h\}} Smooth_{L1} (t^u_i - v_i)
\end{equation}
where 
\begin{displaymath}
\textrm{$Smooth_{L1} (x)$} = \left\{ \begin{array}{ll}
\textrm{$0.5x^2$} & \textrm{if $|x| <1$}\\
\textrm{$|x-0.5|$} & \textrm{otherwise}
\end{array} \right.
\end{displaymath}

The $L_{aff}(m,s)$ is the multinomial cross entropy loss for the affordance detection branch and is computed as follows:

\begin{equation}
L_{aff}(m,s) = \frac{-1}{N}\sum_{i\in RoI} log(m^i_{s_i})  
\end{equation}
where $m^i_{s_i}$ is the softmax output at pixel $i$ for the true label $s_i$; 
%where $s_i$ is the value at pixel $i$ in the target affordance mask $s$ (i.e., the groundtruth affordance label for pixel $i$); 
$N$ is the number of pixels  in the RoI.

In Equation (\ref{eq:loss_fine}), $I[u \ge 1]$ is an indicator function which outputs 1 when $u\ge 1$ and $0$ otherwise. This means that we only define the box location loss $L_{loc}$ and the affordance detection loss $L_{aff}$ only on the positive RoIs. While the object classification loss $L_{cls}$ is defined on both positive and negative RoIs. %from RPN. 

It is worth noting that our loss for affordance detection branch is different from the instance segmentation loss in~ \citep{Kaiming17_MaskRCNN_short, Li2016_FCIS}. In those works, the authors rely on the output of the classification layer to determine the object label. Hence the segmentation in each RoI can be considered as a binary segmentation, i.e., foreground and background. Thus, the authors use per-pixel $sigmoid$ layer and binary cross entropy loss. In our affordance detection problem, the affordance labels are different from the object labels. Furthermore, the number of affordances in each RoI is not binary, i.e., it is always bigger than 2 (including the background). Hence, we rely on a per-pixel $softmax$ and a multinomial cross entropy loss.

\subsection{Training and Inference}

We train the network in an end-to-end manner using stochastic gradient descent with $0.9$ momentum and $0.0005$ weight decay. The network is trained on a Titan X GPU for $200k$ iterations. The learning rate is set to $0.001$ for the first $150k$ and decreased by $10$ for the last $50k$. The input images are resized such that the shorter edge is $600$ pixels, but the longer edge does not exceed $1000$ pixels. In case the longer edge exceeds $1000$ pixels, the longer edge is set to $1000$ pixels, and the images are resized based on this edge. Similar to~\citep{Kaiming17_MaskRCNN_short}, we use $15$ anchors in the RPN ($5$ scales and $3$ aspect ratios). Top $2000$ RoIs from RPN (with a ratio of 1:3 of positive to negative) are subsequently used for computing the multi-task loss. An RoI is considered positive if it has IoU with a groundtruth box of at least 0.5 and negative otherwise. 

During the inference phase, we select the top $1000$ RoIs produced by the RPN and run the object detection branch on these RoIs, followed by a non-maximum suppression~\citep{Shaoqing2015}. From the outputs of the detection branch, we select the outputted boxes that have the classification score higher than $0.9$ as the final detected objects. In case there are no boxes satisfying this condition, we select the one with highest classification score as the only detected object. We use the detected objects as the inputs for affordance detection branch. For each pixel in the detected object, the affordance branch predicts $C+1$ affordance classes. The output affordance label  for each pixel  is achieved by taking the maximum across the affordance classes. Finally, the predicted $244 \times 244$ affordance mask of each object is resized to the object (box) size using the resizing strategy in Section~\ref{subsub:resizing}. 

In case there is the overlap between detected objects, the final affordance label is decided based on the affordance priority. For example, the ``contain" affordance is considered to have low priority than other affordances since there may have other objects laid on it.

\newpage 
\section{Experiments} \label{Sec:exp}

\subsection{Dataset}

\subsubsection{UMD Dataset} The UMD dataset~\citep{Myers15} contains around $30,000$ RGB-D images of daily kitchen, workshop, and garden objects. The RGB-D images of this dataset were captured from a Kinect camera on a rotating table in a clutter-free setup. This dataset has $7$ affordance classes and $17$ object categories. Since there is no groundtruth for the object bounding boxes, we compute the rectangle coordinates of object bounding boxes based on the affordance masks. We follow the split in~\citep{Myers15} to train and test our network.

\subsubsection{IIT-AFF Dataset} 

The work in~\citep{Myers15} proposed the first affordance dataset with pixel-wise labels. The data were collected using a Kinect sensor, which records RGB-D images at a $480 \times 640$ resolution. Although this dataset contains a large amount of annotated images, most of them were captured on a turntable table in a cluttered-free setup. Consequently, the use of this dataset may not be sufficient for robotic applications in real-world cluttered scenes.

\textbf{Data collection} Since CNN requires a large amount of data for training, we introduce a new affordance dataset to fulfill this purpose. In general, we want to create a large-scale dataset that enables the robot to infer properly in real-world scenes after the training step. In order to do this, we first choose a subset of object categories from the ImageNet dataset~\citep{Olga2015}. In addition, we also collect RGB-D images from various cluttered scene setups using an Asus Xtion sensor and a MultiSense-SL camera. The images from the Asus Xtion and MultiSense camera were collected at  $480 \times 640$ and $1024 \times 1024$ resolutions, respectively.

\textbf{Data annotation}  Our dataset provides both bounding box annotations for object detection and pixel-wise labels for affordance detection. We reuse the bounding boxes that come with the images from the ImageNet dataset, while all images are manually annotated with the affordance labels at pixel-level. Since the images from the ImageNet dataset don't have the associated depth maps, we use the state-of-the-art method in~\citep{Fayao2015} to generate the relative depth maps for these images, which can be used by algorithms that need them.

\begin{figure}[ht] 
    \centering

	\includegraphics[width=0.99\linewidth, height=0.78\linewidth]{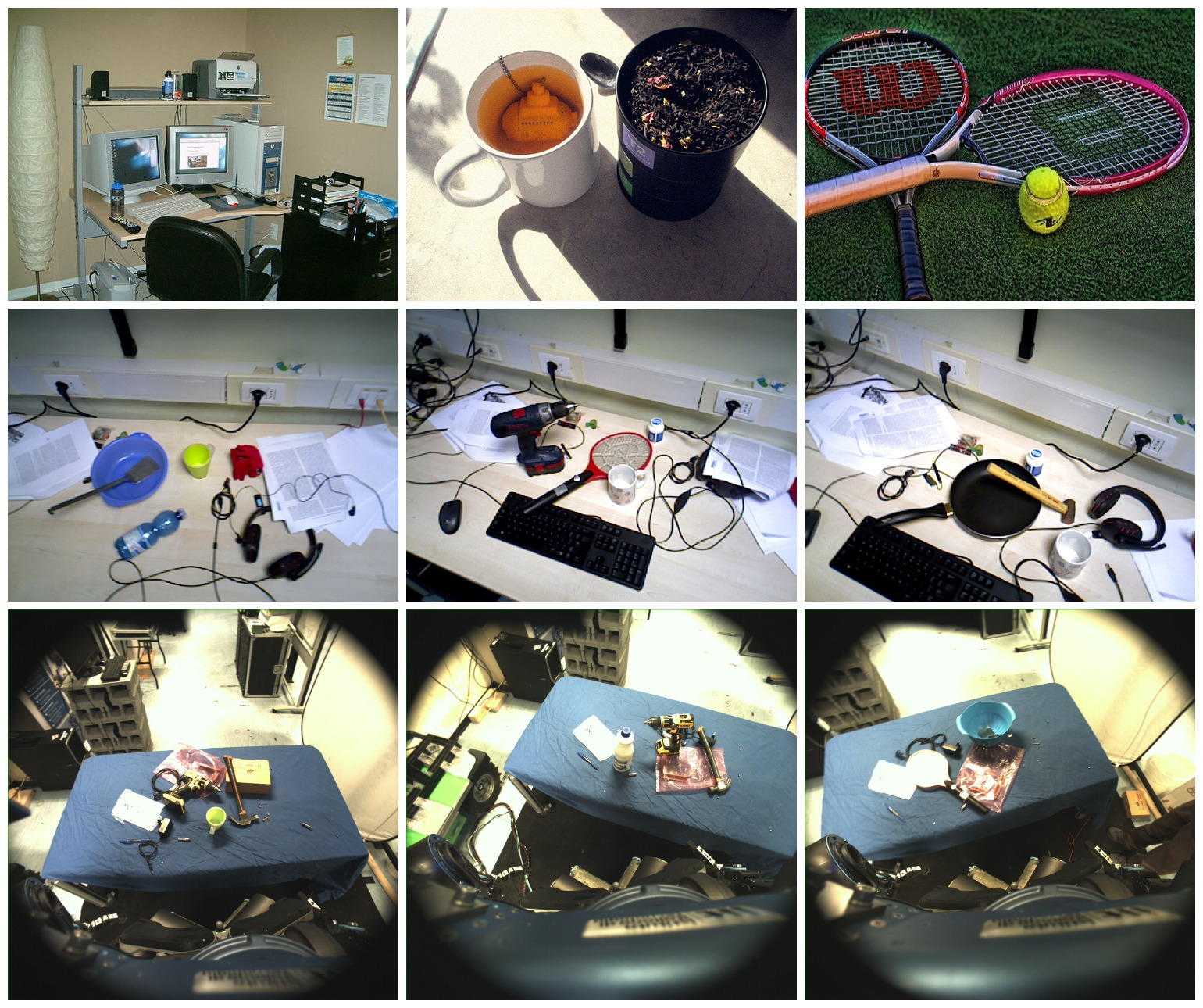}    %\includegraphics[width=0.99\linewidth, height=0.2\linewidth]{example-image-a} 
    \vspace{0.2ex}
    \caption{Example images from our IIT-AFF dataset. \textbf{Top row:} Images from the ImageNet dataset. \textbf{Middle row:} Images from the Asus Xtion camera. \textbf{Bottom row:} Images from the MultiSense-SL camera.}
    \label{Fig:dataset_example_images} 
\end{figure}

\begin{figure}[ht]
  \centering
    \subfigure[Object distribution]{\label{fig:a}\includegraphics[width=0.48\linewidth, height=0.35\linewidth]{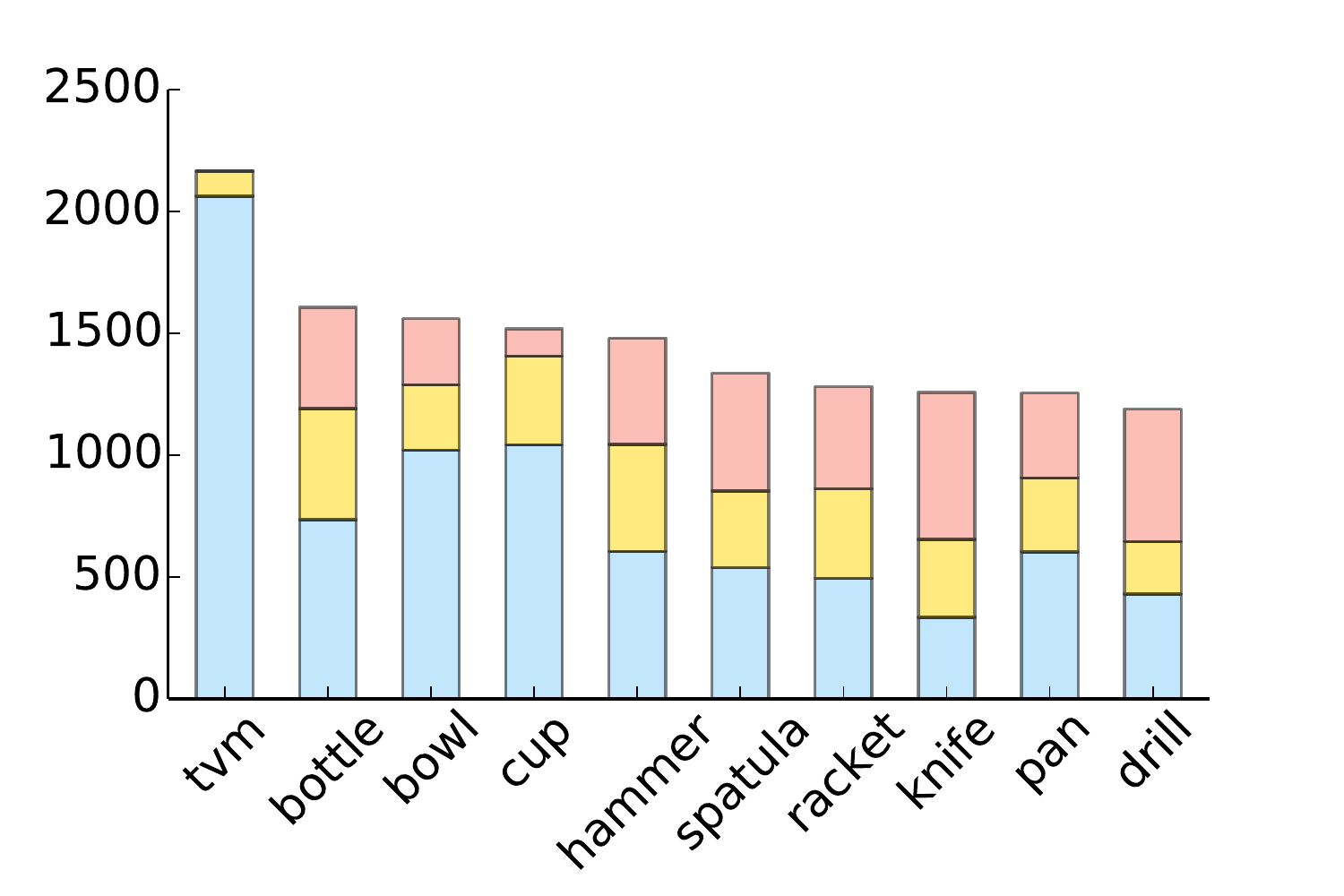}}
    \subfigure[Affordance distribution]{\label{fig:b}\includegraphics[width=0.48\linewidth, height=0.35\linewidth]{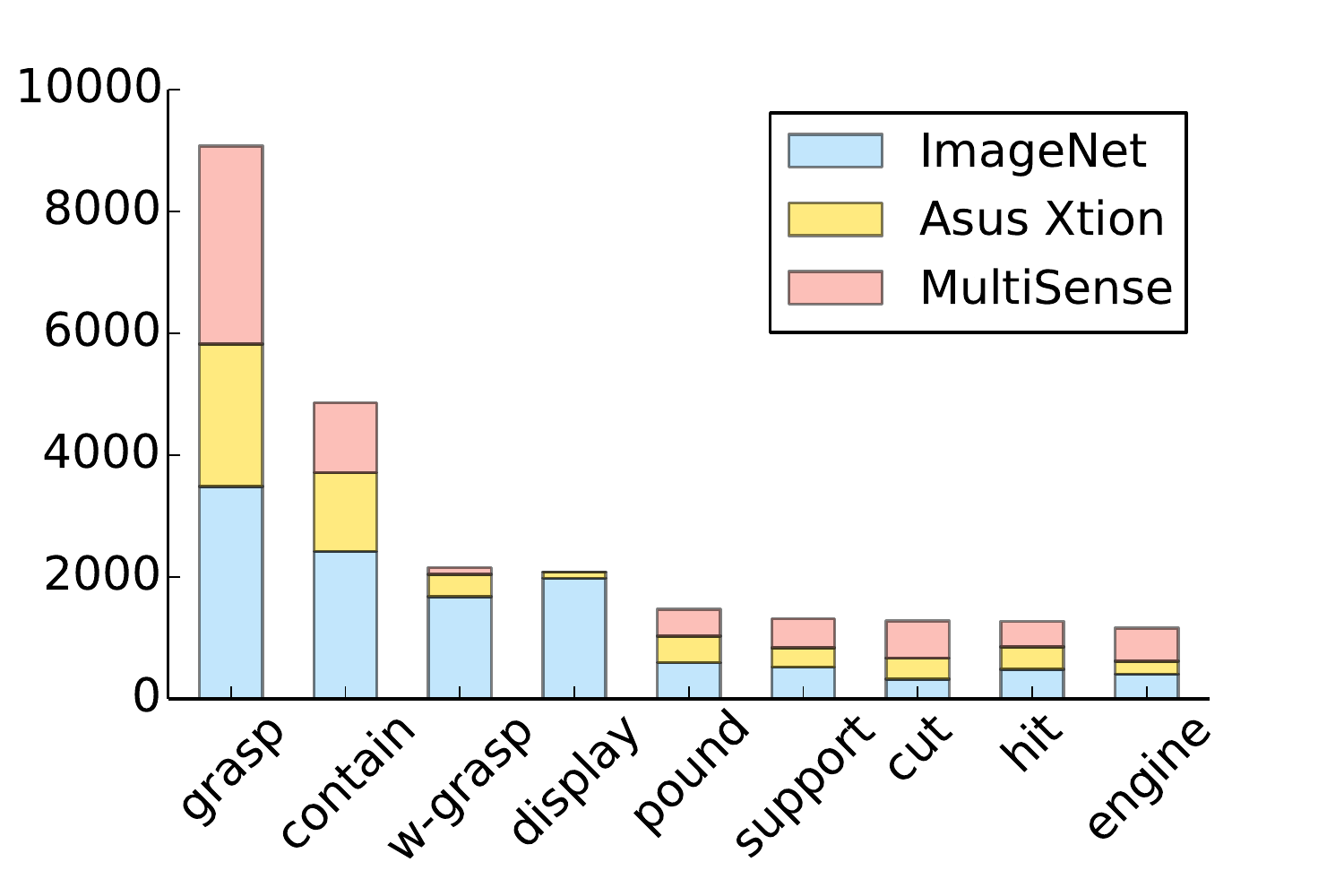}}
     
 \vspace{3.0ex}
 \caption{The statistics of our IIT-AFF dataset. (a) Object distribution as the number of bounding boxes in each object category. (b) Affordance distribution as the number of regions in each affordance class.}
 \label{Fig:dataset_statistic}
\end{figure}

\textbf{Dataset statistic} In particular, our dataset has $10$ object categories (\textsl{bottle}, \textsl{bowl}, \textsl{cup}, \textsl{drill}, \textsl{hammer}, \textsl{knife}, \textsl{monitor}, \textsl{pan}, \textsl{racket}, \textsl{spatula}) and $9$ affordance classes (\texttt{contain}, \texttt{cut}, \texttt{display}, \texttt{engine}, \texttt{grasp}, \texttt{hit}, \texttt{pound}, \texttt{support}, \texttt{w-grasp}), which are common human tools with their related manipulation capabilities (Table~\ref{tb_aff_meaning}). The dataset has $8,835$ images, containing $14,642$ bounding box annotated objects (in which $ 7,866$ bounding boxes come from the ImageNet dataset) and $24,677$ affordance parts of the objects. We use $50\%$ of the dataset for training, $20\%$ for validation, and the rest $30\%$ for inference. Fig.~\ref{Fig:dataset_example_images} and~\ref{Fig:dataset_statistic} show some example images and the statistics of our dataset.

\begin{table}[h!]
\centering\ra{1.3}
%\captionsetup{font=footnotesize, labelfont=bf, textfont=normalfont, justification=centering}
\renewcommand\tabcolsep{1.8pt}
\caption{Description of Object Affordance Labels}
\vspace{2em}
\label{tb_aff_meaning}

\begin{tabular}{ |l||l| } 
\hline
Affordances & Function \\
\hline
\texttt{contain} & Storing/holding liquid/objects (e.g. the inside part of bowls) \\ 
\texttt{cut} & Chopping objects (e.g. the knife blade)  \\ 
\texttt{display} & Showing information (e.g. the monitor screen) \\
\texttt{engine} & Covering engine part of tools (e.g. the drill's engine) \\
\texttt{grasp} & Enclosing by hand for manipulation (e.g. handles of tools)  \\
\texttt{hit} & Striking other objects with refection (e.g. the racket head) \\
\texttt{pound} & Striking other objects with solid part (e.g. the hammer head)  \\
\texttt{support} & Holding other objects with flat surface (e.g. turners) \\
\texttt{w-grasp} & Wrapping by hand for holding (e.g. the outside of a cup) \\
\hline
\end{tabular}
\end{table}

%%%%%%%%%%%%%%%%%%%%%%%%%%%%%
\begin{figure*}[!t]
  \centering
\includegraphics[scale=0.21]{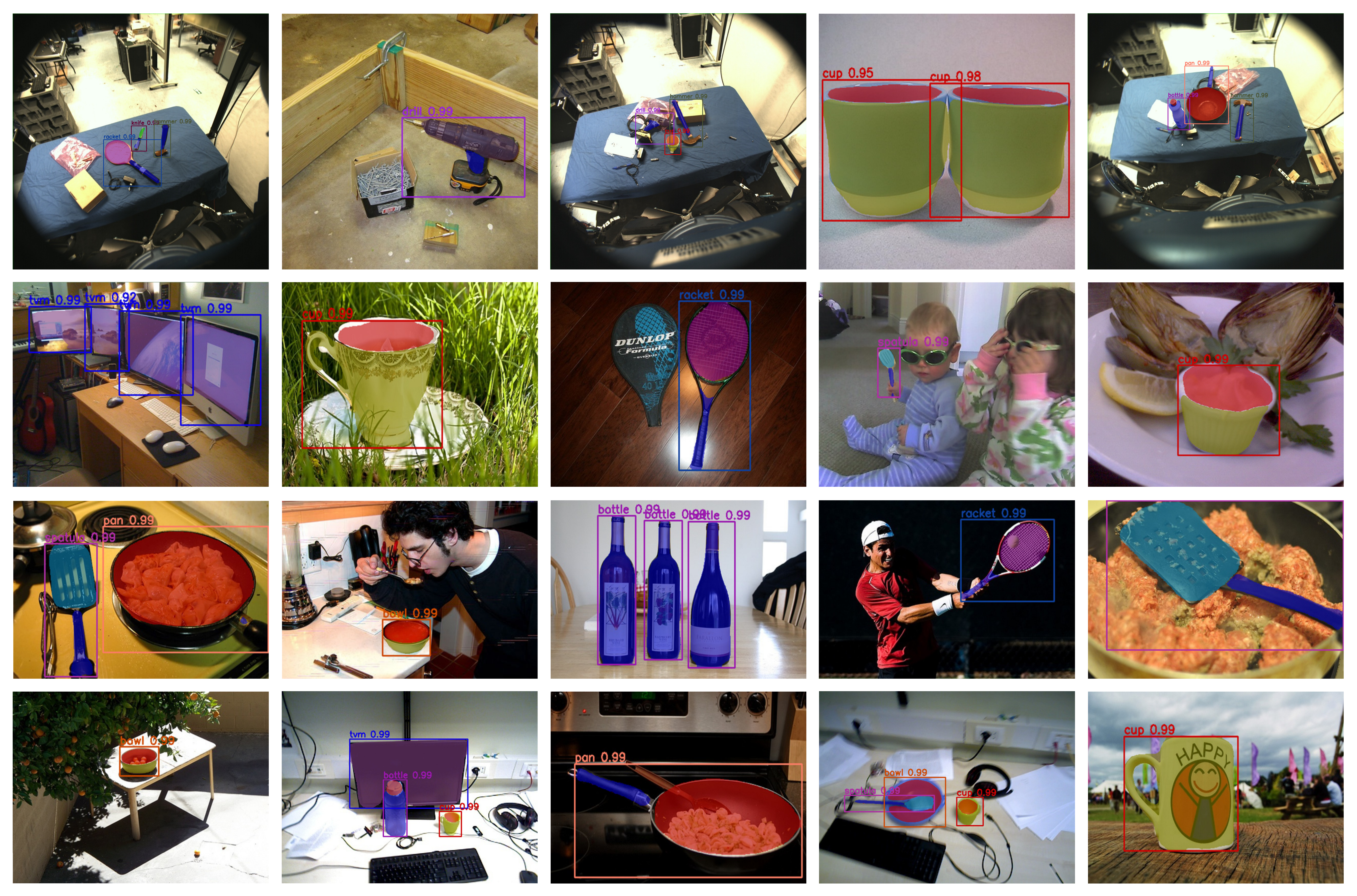}
\vspace{2ex}
 \caption{Examples of affordance detection results by AffordanceNet on the IIT-AFF dataset.}
 \label{Fig:result_aff_detection}
\end{figure*}
%%%%%%%%%%%%%%%%%%%%%%%%%%%%%%% 

\newpage

\subsection{Evaluation Metric and Baseline}
As the standard practice, we compare our affordance results with other methods using the $F_\beta ^w$ metric~\citep{Margolin14}. This metric extends the $F_\beta$ measure and is calculated as follows:
\begin{equation}
	F_\beta ^w = (1 + {\beta ^2})\frac{{Precision ^w \cdot Recall ^w}}{{{\beta ^2} \cdot Precision ^w + Recall ^w}}
\end{equation} 
where $\beta=1$, $Precision ^w$, and $Recall ^w$ are the weighted versions of the standard $Precision$ and $Recall$ measures. The novelty of this measure as explained in~\citep{Margolin14} is to weigh the errors of the pixels by taking into account their location and neighborhood information to overcome three flawed assumptions: interpolation, dependency and equal importance of the prediction map.

We compare our results with the following state-of-the-art approaches: DeepLab~\citep{Chen2016_deeplab} with and without post processing with CRF (denoted as DeepLab and DeepLab-CRF), CNN with encoder-decoder architecture~\citep{Nguyen2016_Aff} on RGB and RGB-D images (denoted as ED-RGB and ED-RGBD), CNN with object detector (BB-CNN) and CRF (BB-CNN-CRF)~\citep{Nguyen2017_Aff}. For the UMD dataset, we also report the results from the geometric features-based approach (HMD and SRF)~\citep{Myers15} and a deep learning-based approach that used both RGB and depth images as inputs (ED-RGBHHA)~\citep{Nguyen2016_Aff}. Note that, all the deep learning-based methods use the VGG16 as the main backbone for a fair comparison.

\subsection{Results}

%%%%%%%%%%%%%%%%%%%%%%%%%%%%%%%%%%%%%%%%%%%

\begin{table}[h]
\centering\ra{1.3}
\caption{Performance on IIT-AFF Dataset}
\renewcommand\tabcolsep{4.0pt}
\label{tb_result_iit}
\hspace{2ex}

\begin{tabular}{@{}rcccccccc@{}}
\toprule 					 &

\multirow{1}{*}[2.5pt]{\scriptsize ED-RGB} & 
\multirow{1}{*}[2.5pt]{\scriptsize ED-RGBD} & 
\multirow{1}{*}[2.5pt]{\scriptsize DeepLab} & 
{\shortstack{\scriptsize DeepLab-\\ \scriptsize CRF}}  & 
\multirow{1}{*}[2.5pt]{\scriptsize BB-CNN} & 
{\shortstack{\scriptsize BB-CNN-\\ \scriptsize CRF}} &
\multirow{1}{*}[2.5pt]{\scriptsize AffordanceNet}  &  \\

%{\shortstack{\ssmall ED-RGB\\ \ssmall ~\citep{Nguyen2016_Aff}} } & 
%{\shortstack{\ssmall ED-RGBD\\ \ssmall ~\citep{Nguyen2016_Aff}}} & 
%%\multirow{1}{*}[2.5pt]{\scriptsize DeepLab~\cite{Chen2016_deeplab}} & 
%{\shortstack{\ssmall DeepLab\\ \ssmall~\citep{Chen2016_deeplab}}} &
%{\shortstack{\ssmall DeepLab-\\ \ssmall CRF~\citep{Chen2016_deeplab}}}  & 
%%\multirow{1}{*}[2.5pt]{\scriptsize BB-CNN} & 
%{\shortstack{\ssmall BB-CNN\\ \ssmall ~\citep{Nguyen2017_Aff}}} &
%{\shortstack{\ssmall BB-CNN-\\ \ssmall CRF~\citep{Nguyen2017_Aff}}} &
%{\shortstack{\ssmall AffordanceNet}} &  \\

\midrule
\texttt{contain} 				& 66.38   & 66.00   & 68.84	& 69.68	& 75.60     & 75.84   & \textbf{79.61} \\
\texttt{cut}					& 60.66   & 60.20   & 55.23	& 56.39	& 69.87     & 71.95   & \textbf{75.68} \\
\texttt{display}				& 55.38   & 55.11   & 61.00	& 62.63 & 72.04     & 73.68   & \textbf{77.81} \\
\texttt{engine} 				& 56.29   & 56.04   & 63.05	& 65.11	& 72.84     & 74.36   & \textbf{77.50} \\
\texttt{grasp}					& 58.96   & 58.59   & 54.31	& 56.24	& 63.72     & 64.26   & \textbf{68.48} \\
\texttt{hit} 					& 60.81   & 60.47   & 58.43	& 60.17	& 66.56     & 67.07   & \textbf{70.75} \\
\texttt{pound} 					& 54.26   & 54.01   & 54.25	& 55.45	& 64.11     & 64.86   & \textbf{69.57} \\
\texttt{support} 				& 55.38   & 55.08   & 54.28	& 55.62	& 65.01     & 66.12   & \textbf{69.81} \\
\texttt{w-grasp} 				& 50.66   & 50.42   & 56.01	& 57.47	& 67.34     & 68.41   & \textbf{70.98} \\
\cline{1-8}
\textbf{Average}				& 57.64   & 57.32   & 58.38	& 59.86	& 68.57     & 69.62   & \textbf{73.35} \\
\bottomrule
\end{tabular}
\end{table}

%%%%%%%%%%%%%%%%%%%%%%%%%%%

\textbf{IIT-AFF Dataset} Table~\ref{tb_result_iit} summarizes results on the IIT-AFF dataset. %From this table, we notice that our AffordanceNet clearly improves the results over the-state-of-the-art methods. 
The results clearly show that AffordanceNet significantly improves over the state of the art. In particular, AffordanceNet boosts the $F_\beta ^w$ score to $73.35$, which is  $3.7\%$ improvement over the second best BB-CNN-CRF. It is worth noting that AffordanceNet achieves this result using an end-to-end architecture, and no further post processing step such as CRF is used. Our AffordanceNet also achieves the best results for all $9$ affordance classes. We also found that for the dataset containing cluttered scenes such as IIT-AFF, the approaches that combine the object detectors with deep networks to predict the affordances (AffordanceNet, BB-CNN) significantly outperform over the methods that use deep networks alone (DeepLab, ED-RGB).

\begin{table}[h]
\centering\ra{1.3}
\renewcommand\tabcolsep{4.0pt}
\caption{Performance on UMD Dataset}
\label{tb_result_umd}
\hspace{2ex}
\begin{tabular}{@{}rcccccccc@{}}
\toprule 			&

\multirow{1}{*}[2.5pt]{\scriptsize HMP} &
\multirow{1}{*}[2.5pt]{\scriptsize SRF} &
\multirow{1}{*}[2.5pt]{\scriptsize DeepLab} &
\multirow{1}{*}[2.5pt]{\scriptsize ED-RGB} &
\multirow{1}{*}[2.5pt]{\scriptsize ED-RGBD} &
\shortstack{\scriptsize ED-RGB\\ \scriptsize HHA}  &
\multirow{1}{*}[2.5pt]{\scriptsize AffordanceNet} &

%& \shortstack{\ssmall HMP\\ \ssmall ~\citep{Myers15}}   
%& \shortstack{\ssmall SRF\\ \ssmall ~\citep{Myers15}}  
%& \shortstack{\ssmall DeepLab\\ \ssmall ~\citep{Chen2016_deeplab}}    
%& \shortstack{\ssmall ED-RGB\\ \ssmall ~\citep{Nguyen2016_Aff}}    
%& \shortstack{\ssmall ED-RGBD\\ \ssmall ~\citep{Nguyen2016_Aff}}     
%& \shortstack{\ssmall ED-RGB\\ \ssmall HHA~\citep{Nguyen2016_Aff}}   
%& \shortstack{\ssmall AffordanceNet}   
\\
\midrule
\texttt{grasp} 				& 0.367   & 0.314   & 0.620				& 0.719	    		& 0.714     & 0.673  & \textbf{0.731} \\
\texttt{w-grasp}			& 0.373   & 0.285   & 0.730				& 0.769  		    & 0.767     & 0.652  & \textbf{0.814} \\
\texttt{cut}				& 0.415   & 0.412   & 0.600				& 0.737   			& 0.723     & 0.685  & \textbf{0.762} \\
\texttt{contain}			& 0.810   & 0.635   & \textbf{0.900}	& 0.817   			& 0.819     & 0.716  & 0.833  \\
\texttt{support} 			& 0.643   & 0.429   & 0.600				& 0.780   			& 0.803     & 0.663  & \textbf{0.821} \\
\texttt{scoop} 				& 0.524   & 0.481  	& \textbf{0.800}	& 0.744  			& 0.757     & 0.635  & 0.793  \\
\texttt{pound} 				& 0.767   & 0.666   & \textbf{0.880}	& 0.794   			& 0.806     & 0.701  & 0.836  \\
\cline{1-8}	
\textbf{Average}			& 0.557   & 0.460   & 0.733				& 0.766   			& 0.770     & 0.675  & \textbf{0.799} \\		
\bottomrule
\end{tabular}
\end{table}
\textbf{UMD Dataset} Table~\ref{tb_result_umd} summarizes results on the UMD dataset. On the average, our AffordanceNet also achieves the highest results on this dataset, i.e., it outperforms the second best (ED-RGBD) $2.9\%$. It is worth noting that the UMD dataset only contains clutter-free scenes, therefore the improvement of AffordanceNet over compared methods is not as high as the one in the real-world IIT-AFF dataset.
% there is no significant improvement between the results of AffordanceNet and other methods that do not use the object detector. Despite this, AffordanceNet still improves the overall performance and achieves the highest $F_\beta ^w$ score in $4$ affordance classes. 
We recall that the AffordanceNet is trained using the RGB images only, while the second best (ED-RGBD) uses both RGB and the depth images. The Table~\ref{tb_result_umd} also clearly shows that the deep learning-based approaches such as AffordanceNet, DeepLab, ED-RGB significantly outperform the hand-designed geometric feature-based approaches (HMP, SRF).

To conclude, our AffordanceNet significantly improves over the state of the art, while it  does not require any extra post processing or data augmentation step. From the robotic point of view, AffordanceNet can be used in many tasks since it provides all the object locations, object categories, and object affordances in an end-to-end manner. The running time of AffordanceNet is around $150ms$ per image on a Titan X GPU, making it is suitable for robotic applications. Our implementation is based on Caffe deep learning library~\citep{Jia14}. The source code and trained models that allow reproducing the results in this paper will be released upon acceptance.

\subsection{Ablation Studies}
\textbf{Effect of Affordance Map Size}\label{Sec:exp_effect_mask_size}
In this section, we analyze the effect of the affordance map size. Follow the setup in Mask RCNN, we use only one deconvolutional layer with parameters ($d=1$, $s=2$, $S_f=4$) to create $14 \times 14$ affordance map from the $7 \times 7$ feature map (denoted as AffordanceNet14). Similarly, we change the parameters to ($d=1$, $s=4$, $S_f=6$) to create the $28 \times 28$ affordance map (denoted as AffordanceNet28). Furthermore, we also setup networks which use two deconvolutional layers to create $56 \times 56$ affordance map (denoted as AffordanceNet56), and three deconvolutional layers to create $112 \times 112$ affordance map (denoted as AffordanceNet112). Finally, to check the effect of the convolutional layers, we also setup a network with $6$ convolutional layers (together with ReLu), follow by a deconvolutional layer that upsampling the $7 \times 7$ feature map to $14 \times 14$ (denoted as AffNet14\_6Conv).

%%%%%%%%%%%%%%%%%%%%%%%%%%%%%%%%%%%
%\vspace{-0.3cm}
\begin{figure*}
\centering
\footnotesize
  \stackunder[7pt]{\includegraphics[width=2.4cm, height=2.25cm]{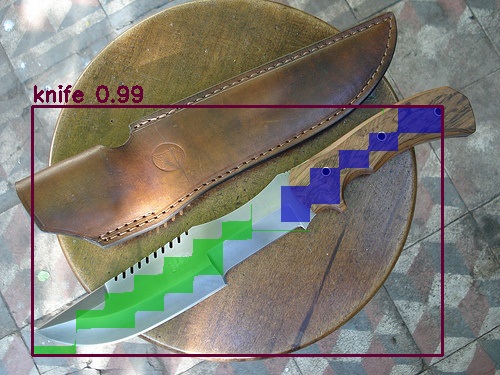}} {\scriptsize {AffordanceNet14}}
  \stackunder[7pt]{\includegraphics[width=2.4cm, height=2.25cm]{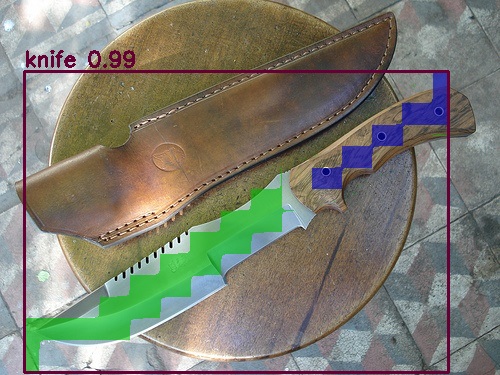}} {\scriptsize AffNet14\_6conv} %\hspace{0.25cm}%
  \stackunder[7pt]{\includegraphics[width=2.4cm, height=2.25cm]{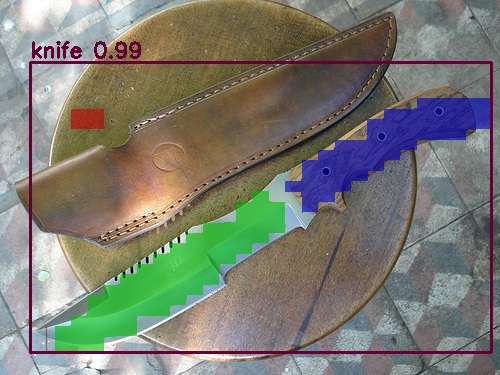}} {\scriptsize AffordanceNet28} %\hspace{-0.25cm}%
  %\hspace{1ex}
  \stackunder[7pt]{\includegraphics[width=2.4cm, height=2.25cm]{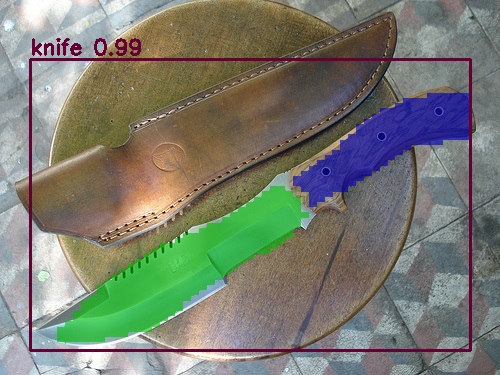}} {\scriptsize AffordanceNet56}
  \stackunder[7pt]{\includegraphics[width=2.4cm, height=2.25cm]{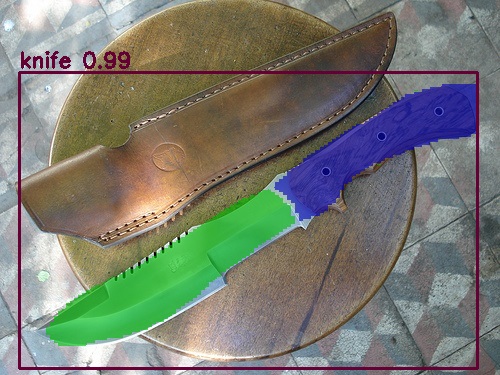}} {\scriptsize AffordanceNet112}
  \stackunder[7pt]{\includegraphics[width=2.4cm, height=2.25cm]{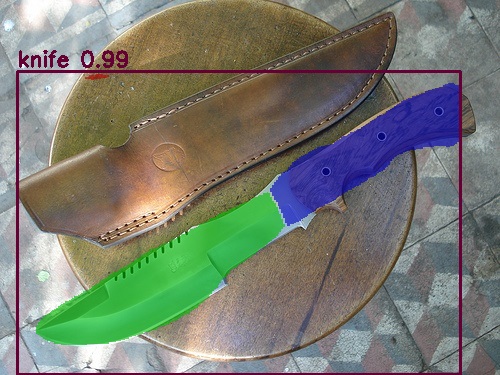}} {\scriptsize AffordanceNet244}
   %\hspace{-0.25cm}%
\vspace{2ex}
\caption{Examples of predicted affordance masks using different mask sizes. The predicted mask is smoother and finer when a bigger mask size is used.}

\label{Fig:result_mask_size_effect} 
\end{figure*}
\vspace{0.5ex}
%%%%%%%%%%%%%%%%%%%%%%%%%%%%%%%%%%%

\begin{figure}
  \centering
  
\subfigure[]{\label{fig_wild_a}\includegraphics[width=0.32\linewidth, height=0.22\linewidth]{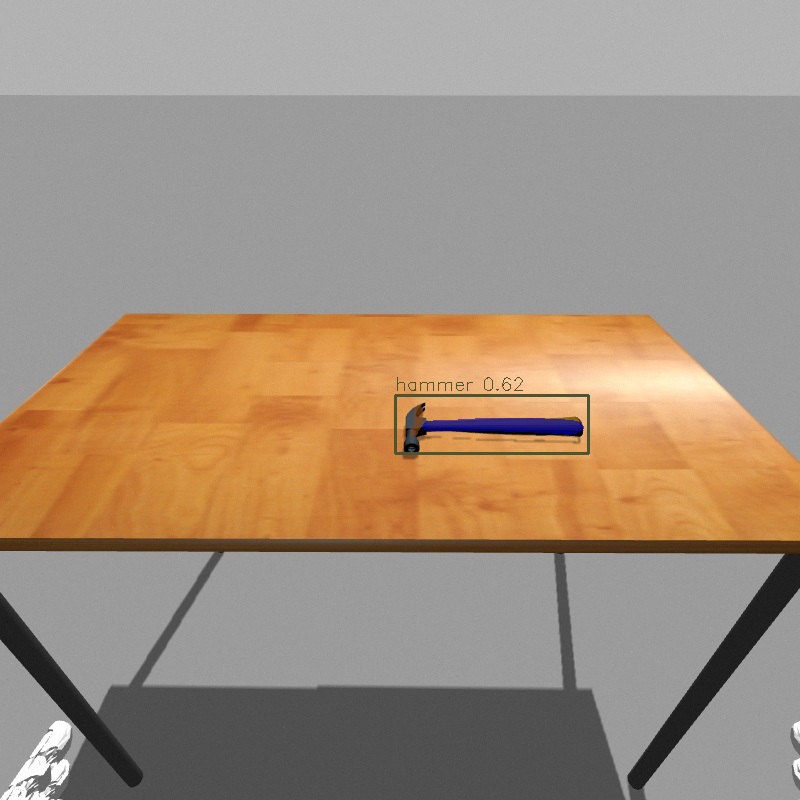}}
\subfigure[]{\label{fig_wild_b}\includegraphics[width=0.32\linewidth, height=0.22\linewidth]{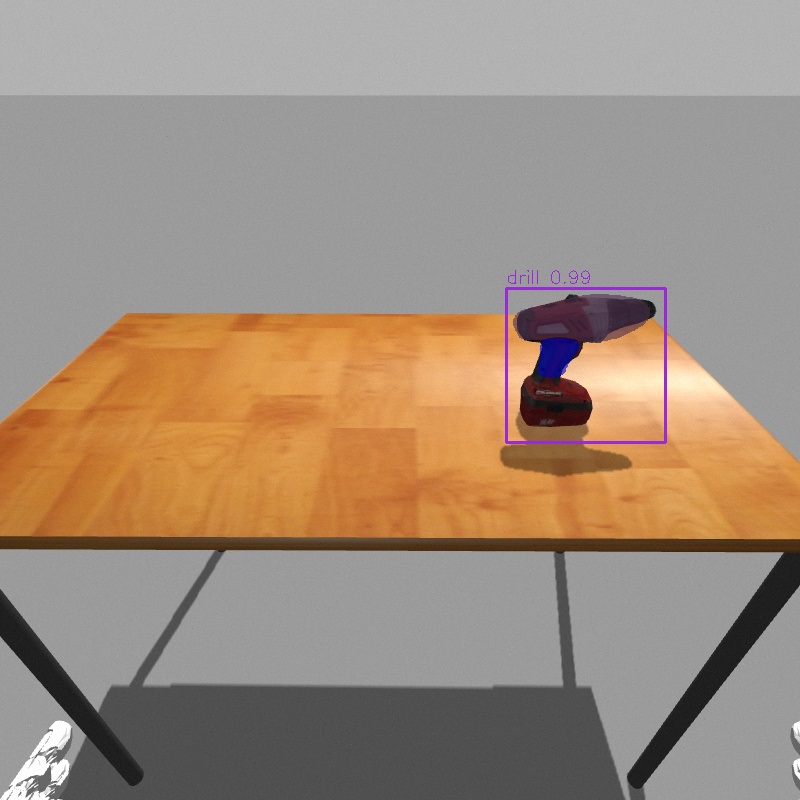}}
\subfigure[]{\label{fig_wild_c}\includegraphics[width=0.32\linewidth, height=0.22\linewidth]{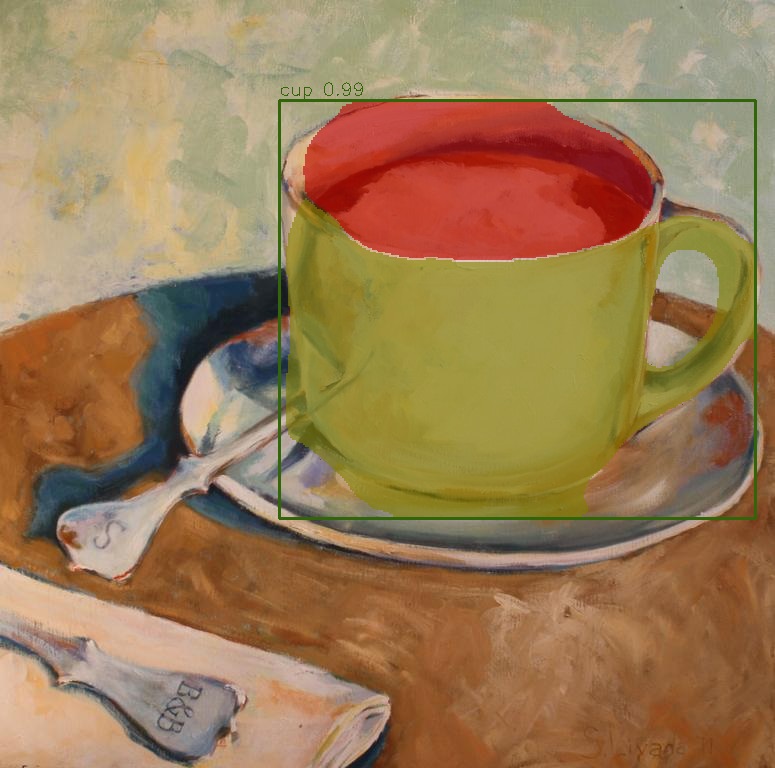}}

	\vspace{2ex}
    
 \caption{Affordance detection in the wild. \textbf{(a)} and \textbf{(b)}: We use AffordanceNet to detect the objects in Gazebo simulation. \textbf{(c)} AffordanceNet also performs well when the input is an artwork.}
 \label{Fig:result_aff_wild}
 \vspace{2ex}
\end{figure}

\vspace{0.3cm}
\begin{table}[!h]
\centering\ra{1.3}
\renewcommand\tabcolsep{8.0pt}
\caption{Effect of Mask Size}
\label{tb_result_masksize_effect}
\hspace{2ex}

\begin{tabular}{@{}rcccccccc@{}}
\toprule 					
%& \shortstack{\ssmall HMP\\ \ssmall ~\cite{Myers15}}   
%& \shortstack{\ssmall SRF\\ \ssmall ~\cite{Myers15}}  
%& \shortstack{\ssmall DeepLab\\ \ssmall ~\cite{Chen2016_deeplab}}    
%& \shortstack{\ssmall ED-RGB\\ \ssmall ~\cite{Nguyen2016_Aff}}    
%& \shortstack{\ssmall ED-RGBD\\ \ssmall ~\cite{Nguyen2016_Aff}}     
%& \shortstack{\ssmall ED-RGB\\ \ssmall HHA~\cite{Nguyen2016_Aff}}   
%& \shortstack{\ssmall E2E-CNN\\ \ssmall (ours)}   
& $F_\beta ^w$
\\
\midrule
AffordanceNet14 				& 57.71    \\
AffordanceNet28					& 66.13    \\
AffordanceNet56					& 71.54    \\
AffordanceNet112				& 72.52    \\ 
AffordanceNet14\_6Conv 			& 60.27    \\
%AffordanceNet244\_$alpha=0.4$ 	& 00.00    \\
\cline{1-2}	
AffordanceNet 					& 73.35    \\
 				
\bottomrule
\end{tabular}
%\vspace{0.3cm}
\end{table}

Table~\ref{tb_result_masksize_effect} summarizes the average $F_\beta ^w$ score of the aforementioned networks on the IIT-AFF dataset. The results show that the affordance detection accuracy is gradually increasing when the bigger affordance map is used. In particular, the AffordanceNet14 gives very poor results since the map size of $14 \times 14$ is too small to represent multiclass affordances. The accuracy is significantly improved when we use the $28 \times 28$ affordance map. However, the improvement does not linearly increase with the affordance map size, it slows down when the bigger mask sizes are used. Note that using the big affordance map can improve the accuracy, but it also increases the number of parameters of the network. In our work, we choose the $244\times244$ map size for AffordanceNet since it both gives the good accuracy and can be trained with a Titan X GPU. We also found that using more convolutional layers (as in AffordanceNet14\_6Conv) can also improve the accuracy, but it still requires to upsample the affordance map to high resolution in order to achieve good results. Fig~\ref{Fig:result_mask_size_effect} shows some example results when different affordance map sizes are used.

%\vspace{0.2cm}
\textbf{Affordance Detection in The Wild} The experimental results on the simple constrained environment UMD dataset and  the real-world IIT-AFF dataset show that the AffordanceNet performs well on public research datasets. However, real-life images may be more challenging. In this study, we show some qualitative results to demonstrate that the AffordanceNet can generalize well in other testing environments. As illustrated in Fig~\ref{Fig:result_aff_wild}, our AffordanceNet can successfully detect the objects and their affordances from artwork images or images from a simulated camera in Gazebo simulation~\citep{MingoHoffman2014}. Although this result is qualitative, it shows that AffordanceNet is applicable for wide ranges of applications, including in simulation environment which is crucial for developing robotic applications. 

%%%%%%%%%%%%%%%%%%%%%%%%%%%%%%%%%%%%%%%%%%
\begin{figure}
\center
\footnotesize
  \stackunder[5pt]{\includegraphics[width=0.25\linewidth, height=0.3\linewidth]{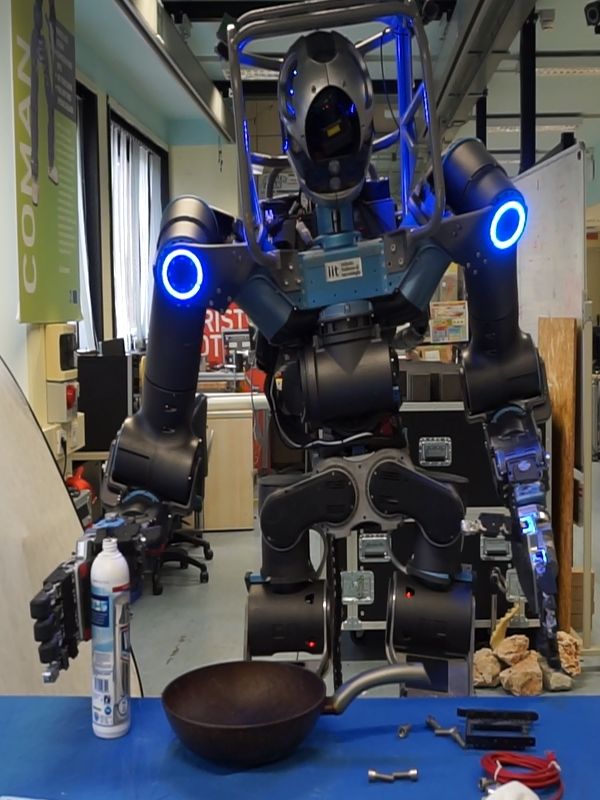}} {}\hspace{0.1cm}%
  \stackunder[5pt]{\includegraphics[width=0.25\linewidth, height=0.3\linewidth]{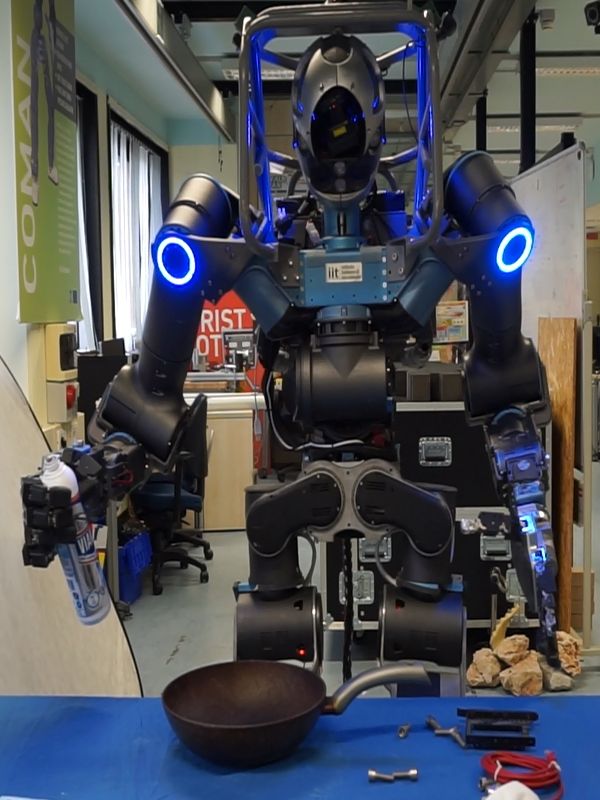}} {}\hspace{0.1cm}%
  \stackunder[5pt]{\includegraphics[width=0.25\linewidth, height=0.3\linewidth]{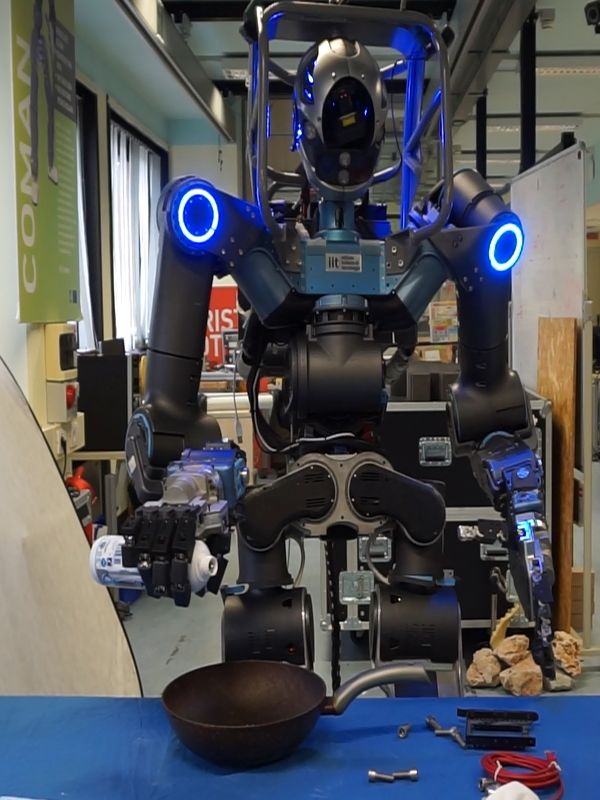}} {}%\hspace{0.1cm}%
\vspace{3ex}
\caption{WALK-MAN is performing a pouring task. % using the visual information from AffordanceNet. 
The outputs of the object detection branch help the robot to recognize and localize the objects (i.e., bottle, pan) while the outputs of the affordance detection branch help the robot to perform the task (i.e., where on the bottle the robot should grasp and where on the pan the water should be poured).}

\label{Fig:robot_pouring} 
%\vspace{0.3cm}
\end{figure}

\newpage

\subsection{Robotic Applications}
Since the AffordanceNet can detect both the objects and their affordances at the speed of $150ms$ per image, 
it is quite suitable for robotic applications. To demonstrate that, we use the humanoid robot WALK-MAN~\citep{Niko2016_full} to perform different manipulation experiments. The robot is controlled in real-time using the XBotCore framework~\citep{muratore2017xbotcore}. The whole-body motion planning is generated by OpenSoT library~\citep{Rocchi15}, while the AffordanceNet is used to provide visual information for the robot. Note that, from the 2D information outputted by AffordanceNet, we use the corresponding depth image to project it into 3D space, to be used in the real robot. Using this setup, the robot can perform different tasks such as grasping, pick-place, and pick-pouring. It is worth noting that all information produced by the AffordanceNet, i.e. the object locations, object labels, and object affordances are very useful for the tasks. 
For example, the robot knows where to grasp a \textsl{bottle} via the bottle's \texttt{grasp} affordance, and where to pour the water into a \textsl{pan} via the pan's \texttt{contain} affordance (see Fig.~\ref{Fig:robot_pouring}). Our experimental video can be found at the following link: {\url{https://sites.google.com/site/affordancenetwork/}}

\chapter{Fine-grained Action Understanding}
\label{ch_act_understanding}

\section{Introduction} \label{Sec:Intro}
While humans can effortlessly understand the actions and imitate the tasks by just \textit{watching} someone else, making the robots to be able to perform actions based on observations of human activities is still a major challenge in robotics~\citep{Chrystopher09}. By understanding human actions, robots may acquire new skills, or perform different tasks, without the need for tedious programming. It is expected that the robots with these abilities will play an increasingly more important role in our society in areas such as assisting or replacing humans in disaster scenarios, taking care of the elderly, or helping people with everyday life tasks. Recently, this problem has been of great interest to researchers, many approaches have been proposed to tackle different tasks such as pouring water~\citep{pastor2009learning}, drawer opening~\citep{rana2017towards}, and multi-stage manipulation~\citep{zhang18_vr_manipulation}.

In this work, we argue that there are two main capabilities that a robot must develop to be able to replicate human activities: \textit{understanding} human actions, and \textit{imitating} them. The imitation step has been widely investigated in robotics within the framework of learning from demonstration (LfD)~\citep{Brenna2009}. In particular, there are two main approaches in LfD that focus on improving the accuracy of the imitation process: kinesthetic teaching~\citep{Akgun2012} and motion capture~\citep{Koenemann2014}. While the first approach needs the users to physically move the robot through the desired trajectories, the second approach uses a bodysuit or camera system to capture human motions. Although both approaches successfully allow a robot to imitate a human, the number of actions that the robot can learn is quite limited due to the need of using expensively physical systems (i.e., real robot, bodysuit, etc.) to capture the training data~\citep{Akgun2012, Koenemann2014}. 

The \textit{understanding} step, on the other hand, receives more attention from the computer vision community. Two popular problems that receive a great deal of interest are video classification~\citep{Karpathy2014} and action recognition~\citep{Simonyan2014}. Recently, with the rise of deep learning, the video captioning task~\citep{Venugopalan2016} has become more feasible to tackle. Unlike the video classification or detection tasks which output only the class identity, the output of the video captioning task is a natural language sentence, which is potentially useful in robotic applications.

While the field of LfD focuses mainly on the imitation step, we focus on the understanding step, but our proposed method also allows the robot to perform useful tasks via the output commands. Our goal is to bridge the gap between computer vision and robotics, by developing a system that helps the robot understand human actions, and use this knowledge to complete useful tasks. Furthermore, since our method provides a meaningful way to let the robots understand human demonstrations by encoding the knowledge in the video, it can be integrated with any LfD techniques to improve the manipulation capabilities of the robot.

In particular, we cast the problem of translating videos to commands as a video captioning task: given a video, the goal is to translate this video to a command. Although we are inspired by the video captioning field, there are two key differences between our approach and the traditional video captioning task: (i) we use the \textit{grammar-free} format in the captions for the convenience in robotic applications, and (ii) we aim at learning the commands through the demonstration videos that contain the \textit{fine-grained} human actions. The use of fine-grained classes forms a more challenging problem than the traditional video captioning task since the fine-grained actions usually happen within a short duration, and there is usually ambiguous information between these actions. To effectively learn the fine-grained actions, unlike the traditional video and image captioning methods~\citep{yao2015describing, you16_image_captioning} that mainly investigate the use of \textit{visual attention} to improve the result, we propose two deep network architectures (S2SNet and V2CNet) that focus on learning and understanding the human action for this problem.

\section{Related Work} \label{Sec:rw}

\textbf{Learning from Demonstration} LfD techniques are widely used in the robotics community to teach the robots new skills based on human demonstrations~\citep{Brenna2009}. Two popular LfD approaches are kinesthetic teaching~\citep{Pastor11, Akgun2012} and sensor-based demonstration~\citep{Calinon09, kruger2010learning}. Recently, ~\cite{Koenemann2014} introduced a method to transfer complex whole-body human motions to a humanoid robot in real-time. ~\cite{Welschehold2016} proposed to transform human demonstrations into hand-object trajectories in order to adapt to robotic manipulation tasks. The advantage of LfD approaches is their abilities to let the robots accurately repeat human motions, however, it is difficult to expand LfD techniques to a large number of tasks since the training process is usually designed for a specific task or needs training data from the real robotic systems~\citep{Akgun2012}.

\textbf{Action Representation} From a computer vision point of view, ~\cite{Aksoy2016} represented the continuous human actions as ``semantic event chains" and solved the problem as an activity detection task. ~\cite{Yang2015} proposed to learn manipulation actions from unconstrained videos using CNN and grammar based parser. However, they need an explicit representation of both the objects and grasping types to generate command sentences. ~\cite{lee2013syntactic} captured the probabilistic activity grammars from the data for imitation learning. ~\cite{Ramirez2015} extracted semantic representations from human activities in order to transfer skills to the robot. Recently, ~\cite{Plappert17} introduced a method to learn bidirectional mapping between human motion and natural language with RNN. In this paper, we propose to directly learn the commands from the demonstration videos without any prior knowledge. Our method takes advantage of CNN to extract robust features, and RNN to model the sequences, while being easily adapted to any human activity.

\textbf{Command Understanding} Currently, commands are widely used to communicate and control real robotic systems. However, they are usually carefully programmed for each task. This limitation means programming is tedious if there are many tasks. To automatically allow the robot to understand the commands, ~\cite{Tellex2011} formed this problem as a probabilistic graphical model based on the semantic structure of the input command. Similarly, ~\cite{Guadarrama13} introduced a semantic parser that used both natural commands and visual concepts to let the robot execute the task. While we retain the concepts of~\citep{Tellex2011, Guadarrama13}, the main difference in our approach is that we directly use the grammar-free commands from the translation module. This allows us to use a simple similarity measure to map each word in the generated command to the real command that can be used on the real robot.

\textbf{Video Captioning} With the rise of deep learning, the video captioning problem becomes more feasible to tackle. ~\cite{Donahue2014} introduced a first deep learning framework that can interpret an input video to a sentence. The visual features were first extracted from the video frames then fed into a LSTM network to generate the video captions. Recently, ~\cite{Haonan2016} introduced a new hierarchical RNN to generate one or multiple sentences to describe a video. ~\cite{Venugopalan2016} proposed a sequence-to-sequence model to generate the video captions from both RGB and optical flow images. The authors in~\citep{yao2015describing,Ramanishka2017cvpr} investigated the use of visual attention mechanism for this problem. Similarly, ~\cite{pan17_semantic_attributes} proposed to learn semantic attributes from the image then incorporated this information into a LSTM network. In this work, we cast the problem of translating videos to commands as a video captioning task to build on the strong state the art in computer vision. However, unlike~\citep{yao2015describing,Ramanishka2017cvpr} that explore the visual attention mechanism to improve the result, we focus on the fine-grained actions in the video. Our hypothesis is based on the fact that the fine-grained action is the key information in the video-to-command task, hence plays a significant contribution to the results.

\textbf{Reinforcement/Meta Learning} Recently, reinforcement learning and meta-learning techniques are also widely used to solve the problem of learning from human demonstrations. ~\cite{Rhinehart17_activity} proposed a method to predict the outcome of the human demonstrations from a scene using inverse reinforcement learning. ~\cite{Stadie17} learned a reward function based on human demonstrations of a given task in order to allow the robots to execute some trials, then maximize the reward to reach the desired outcome. With a different viewpoint, ~\cite{Finn17_oneshot_learning} proposed an one-shot visual imitation learning method to let the robot perform the tasks from just one single demonstration. More recently, ~\cite{yu2018one} presented a new approach for one-shot learning by fusing human and robot demonstration data to build up prior knowledge through meta-learning. The main advantage of meta-learning approach is it allows the robot to replicate human actions from just one or few demonstrations, however handling domain shift and the dependence on the data from the real robotic system are still the major challenges in this approach.

\newpage

\section{Problem Formulation}
In order to effectively generate both the output command and understand the fine-grained action in the video, we cast the problem of translating videos to commands as a video captioning task, and use a TCN network to classify the actions. Both tasks use the same input feature and are jointly trained using a single multi-task loss function. In particular, the input video is considered as a list of frames, presented by a sequence of features  $\mathbf{X} = (\mathbf{x}_1, \mathbf{x}_2, ..., \mathbf{x}_n)$ from each frame. The output command is presented as a sequence of word vectors $\mathbf{Y}^w=(\mathbf{y}_1^w, \mathbf{y}_2^w, ..., \mathbf{y}_m^w)$, in which each vector $\mathbf{y}^w$ represents one word in the dictionary $D$. Each input video also has an action groundtruth label, represented as an one-hot vector $\mathbf{Y}^a \in {\mathbb{R}^{|C|}}$, where $C$ is the number of action classes, to indicate the index of the truth action class. Our goal is to simultaneously find for each sequence feature $\mathbf{X}_i$ its most probable command $\mathbf{Y}_i^w$, and its best suitable action class $\mathbf{Y}_i^a$. In practice, the number of video frames $n$ is usually higher than the number of words $m$. To make the problem more suitable for robotic applications, we use a dataset that contains mainly human demonstrations and assume that the command is in grammar-free format.

\subsection{Command Embedding}
Since a command is a list of words, we have to represent each word as a vector for computation. There are two popular techniques for word representation: \textit{one-hot} encoding and \textit{word2vec}~\citep{Mikolov2013} embedding. Although the one-hot vector is high dimensional and sparse since its dimensionality grows linearly with the number of words in the vocabulary, it is straightforward to use this embedding in the video captioning task. In this work, we choose the one-hot encoding technique as our word representation since the number of words in our dictionary is relatively small. The one-hot vector $\mathbf{y}^w \in {\mathbb{R}^{|D|}}$ is a binary vector with only one non-zero entry indicating the index of the current word in the vocabulary. Formally, each value in the one-hot vector $\mathbf{y}^w$ is defined by:

 \begin{equation}
    \mathbf{y}^w(i)=
    \begin{cases}
      1, & \text{if}\ i=ind(\mathbf{y}^w) \\
      0, & \text{otherwise}
    \end{cases}
  \end{equation}
where $ind(\mathbf{y}^w)$ is the index of the current word in the dictionary $D$. In practice, we add an extra word \textsf{EOC} to the dictionary to indicate the end of command sentences.

\subsection{Visual Features}
As the standard practice in the video captioning task~\citep{Venugopalan2016, Ramanishka2017cvpr}, the visual feature from each video frame is first extracted offline then fed into the captioning module. This step is necessary since we usually have a lot of frames from the input videos. In practice, we first sample $n$ frames from each input video in order to extract deep features from the images. The frames are selected uniformly with the same interval if the video is too long. In case the video is too short and there are not enough $n$ frames, we create an artificial frame from the mean pixel values of the ImageNet dataset~\citep{Olga2015} and pad this frame at the end of the list until it reaches $n$ frames. We then use the state-of-the-art CNN to extract deep features from these input frames. Since the visual features provide the key information for the learning process, three popular CNN are used in our experiments: VGG16~\citep{SimonyanZ14}, Inception\_v3~\citep{Szegedy16_Inception}, and ResNet50~\citep{He2016}.

Specifically, for the VGG16 network, the features are extracted from its last fully connected $\texttt{fc2}$ layer. For the Inception\_v3 network, we extract the features from its $\texttt{pool\_3:0}$ tensor. Finally, we use the features from $\texttt{pool5}$ layer of the ResNet50 network. The dimension of the extracted features is $4096$, $2048$, $2048$, for the VGG16, Inception\_v3, and ResNet50 network, respectively. All these CNN are pretrained on ImageNet dataset for image classifications. We notice that the names of the layers we mention here are based on the Tensorflow~\citep{TensorFlow2015} implementation of these networks.

\section{S2SNet Architecture}\label{seq2seq_arch}
\subsection{Architecture}

\begin{figure*}[!htbp] 
    \centering
	\includegraphics[scale=0.27]{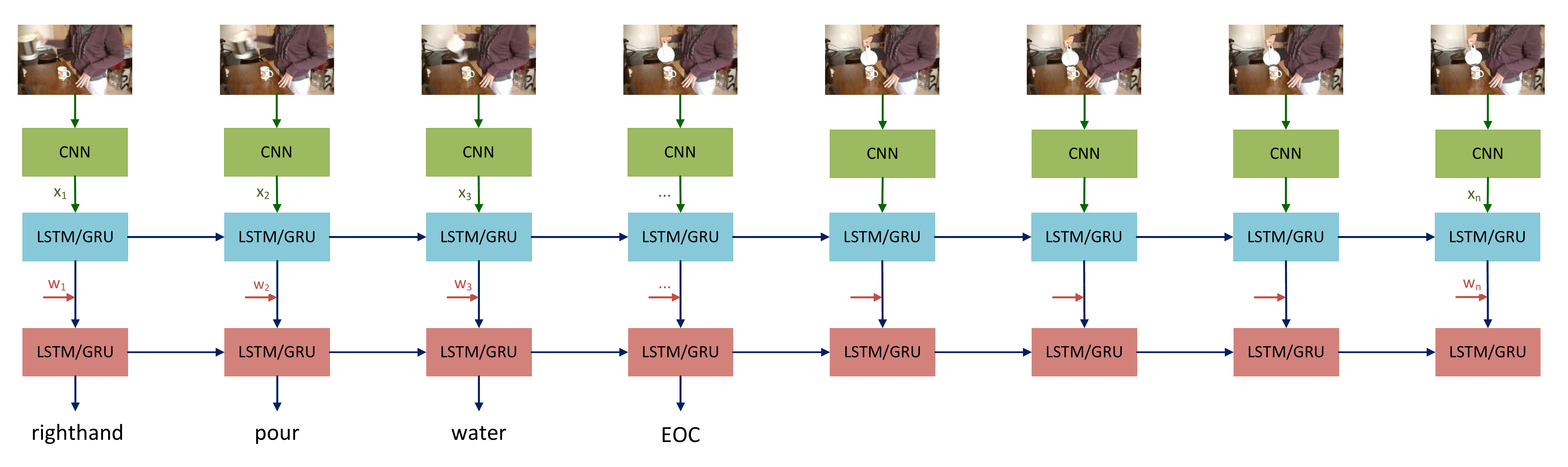} 
    \vspace{1.0ex}
    \caption{An overview of our sequence to sequence approach. We first extract the deep features from the input frames using CNN. Then the first LSTM/GRU layer is used to encode the visual features. The input words are fed to the second LSTM/GRU layer and this layer sequentially generates the output words.}
    \label{Fig:overview_s2s} 
\end{figure*}

Our S2SNet is based on the encoder-decoder scheme~\citep{Venugopalan2016, Ramanishka2017cvpr}, which is adapted from the popular sequence to sequence model~\citep{Sutskever2014_Seq} in machine translation. Although recent approaches to video captioning problem use attention mechanism~\citep{Ramanishka2017cvpr} or hierarchical RNN~\citep{Haonan2016}, our proposal solely relies on the neural architecture. Based on the input data characteristics, our network smoothly encodes the input visual features and generates the output commands, achieving a fair improvement over the state of the art without using any additional modules.

In particular, given an input video, we first extract visual features from the video frames using the pretrained CNN network. These features are encoded in the first RNN layer to create the encoder hidden state. The input words are then fed to the second RNN layer, and this layer will decode sequentially to generate a list of words as the output command. Fig.~\ref{Fig:overview_s2s} shows an overview of our approach. More formally, given an input sequence of features $\mathbf{X} = (\mathbf{x}_1, \mathbf{x}_2, ..., \mathbf{x}_n)$, we want to estimate the conditional probability for an output command  $\mathbf{Y}=(\mathbf{y}_1, \mathbf{y}_2, ..., \mathbf{y}_m)$ as follows:

\begin{equation} \label{Eq_mainP1}
P(\mathbf{y}_1, ..., \mathbf{y}_m | \mathbf{x}_1, ..., \mathbf{x}_n) = \prod\limits_{i = 1}^m {P({\mathbf{y}_i}|{\mathbf{y}_{i - 1}}, ...,{\mathbf{y}_1}},\mathbf{X})
\end{equation}

Since we want a generative model that encodes a sequence of features and produces a sequence of words in order as a command, the LSTM/GRU is well suitable for this task. Another advantage of LSTM/GRU is that they can model the long-term dependencies in the input features and the output words. In practice, we conduct experiments with the LSTM and GRU network as our RNN, while the input visual features are extracted from the VGG16, Inception\_v3, and ResNet50 network, respectively.

In the encoding stage, the first LSTM/GRU layer converts the visual features $\mathbf{X} = (\mathbf{x}_1, \mathbf{x}_2, ..., \mathbf{x}_n)$ to a list of hidden state vectors $\mathbf{H}^e = (\mathbf{h}_1^e, \mathbf{h}_2^e, ..., \mathbf{h}_n^e)$ (using Equation~\ref{Eq_LSTM} for LSTM or Equation~\ref{Eq_GRU} for GRU). Unlike~\citep{venugopalan2014translating} which takes the average of all $n$ hidden state vectors to create a fixed-length vector, we directly use each hidden vector $\mathbf{h}_i^e$ as the input $\mathbf{x}_i^d$ for the second decoder layer. This allows the smooth transaction from the visual features to the output commands without worrying about the harsh average pooling operation, which can lead to the loss of temporal structure underlying the input video.

In the decoding stage, the second LSTM/GRU layer converts the list of hidden encoder vectors $\mathbf{H}^e$ into the sequence of hidden decoder vectors $\mathbf{H}^d$. The final list of predicted words $\mathbf{\hat{Y}}$ is achieved by applying a softmax layer on the output $\mathbf{H}^d$ of the LSTM/GRU decoder layer. In particular, at each time step $t$, the output $\mathbf{z}_t$ of each LSTM/GRU cell in the decoder layer is passed though a linear prediction layer $\hat{\mathbf{y}}=\mathbf{W}_{z}\mathbf{z}_t+\mathbf{b}_z$, and the predicted distribution $P(\mathbf{y}_t)$ is computed by taking the softmax of $\hat{\mathbf{y}_t}$ as follows:

\begin{equation}
P(\mathbf{y}_t=\mathbf{w}|\mathbf{z}_t)=\frac{\text{exp}(\hat{\mathbf{y}}_{t,w})}{\sum_{w'\in{D}}\text{exp}(\hat{\mathbf{y}}_{t,w'})}
\end{equation}
where $\mathbf{W}_z$ and $\mathbf{b}_z$ are learned parameters, $\mathbf{w}$ is a word in the dictionary $D$.

In this way, the LSTM/GRU decoder layer sequentially generates a conditional probability distribution for each word of the output command given the encoded features representation and all the previously generated words. In practice, we preprocess the data so that the number of input words $m$ is equal to the number of input frames $n$. For the input video, this is done by uniformly sampling $n$ frames in the long video, or padding the extra frame if the video is too short. Since the number of words $m$ in the input commands is always smaller than $n$, we pad a special empty word to the list until we have $n$ words.

\subsection{Training}

The network is trained end-to-end with Adam optimizer~\citep{kingma2014adam} using the following objective function:

\begin{equation} \label{Eq_LossFunction}
\mathop {\arg \max }\limits_\theta  \sum\limits_{i = 1}^m {logP({\mathbf{y}_i}|{\mathbf{y}_{i - 1}},...,{\mathbf{y}_1};\theta)} 
\end{equation}
where $\theta$ represents the parameters of the network.

During the training phase, at each time step $t$, the input feature $\mathbf{x}_t$ is fed to an LSTM/GRU cell in the encoder layer along with the previous hidden state $\mathbf{h}_{t-1}^e$ to produce the current hidden state $\mathbf{h}_t^e$. After all the input features are exhausted, the word embedding and the hidden states of the first LSTM/GRU encoder layer are fed to the second LSTM/GRU decoder layer. This decoder layer converts the inputs into a sequence of words by maximizing the log-likelihood of the predicted word (Equation~\ref{Eq_LossFunction}). This decoding process is performed sequentially for each word until the network generates the end-of-command (\textsf{EOC}) token.

\section{V2CNet Architecture}

\begin{figure*}[!htbp] 

    \centering
	\includegraphics[scale=0.262]{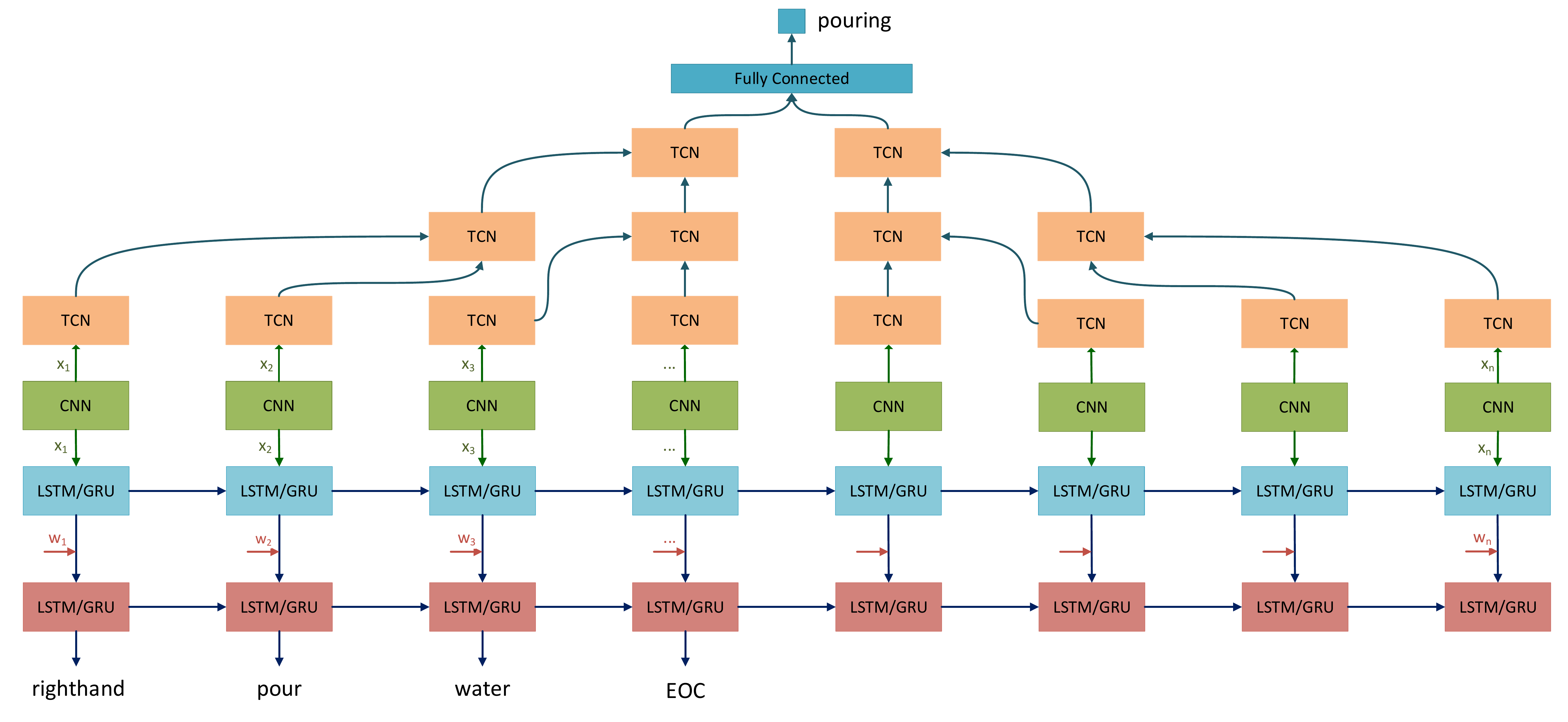} 
    \vspace{2.0ex}
    \caption{A detailed illustration of our V2CNet architecture. The network is composed of two branches: a classification branch and a translation branch. The input for both branches is the visual features extracted by CNN. The classification branch uses the input features to learn the action through a TCN network. The translation branch has the encoder-decoder architecture with two LSTM/GRU layers. The first LSTM/GRU layer is used to encode the visual features, then the input words are combined with the output of the first LSTM/GRU layer and fed into the second LSTM/GRU layer in order to sequentially generate the output words as a command.}
    \label{Fig:v2c_overview} 
\end{figure*}

Since the S2SNet (Section~\ref{seq2seq_arch}) only uses the visual feature, it does not handle the fine-grained actions effectively. To overcome this limitation, we propose to use the Temporal Convolutional Networks (TCN) to explicitly classify the human actions. This TCN network can be considered as an \textit{action attention} mechanism since it does not focus on the visual part of the data, but deeply encodes the fine-grained actions that humans are performing in the videos.

Based on the aforementioned observation, we design a network with two branches: a translation branch that interprets videos to commands using a RNN network, and an action classification branch that classifies the fine-grained human actions with a TCN network. The intuition of our design is that since both the classification and translation branches are jointly trained using a single loss function, the parameters of both branches are updated using the same gradient signal through backpropagation. Therefore, the classification branch will encourage the translation branch to generate the correct fine-grained action, which is the key information to understand human demonstrations. We experimentally demonstrate that by jointly train both branches, the translation results are significantly improved.

To simultaneously generate the command sentence and classify the fine-grained actions, we design a deep network with two branches: a classification branch and a translation branch. The classification branch is a TCN network that handles the human action, while the translation branch is a RNN with encoder-decode architecture to encode the visual features and sequentially generate the output words as a command. Both branches share the input visual features from the video frames and are trained end-to-end together using a single multi-task loss function. Our intuition is that since the translation branch has to generate the command from a huge space of words, it may not effectively encode the fine-grain actions. Therefore, we integrate the classification branch which is trained directly on the smaller space of action classes to ease the learning process. In practice, the parameters of both branches are updated using the same gradient signal, hence the classification branch will encourage the translation branch to generate the correct fine-grained action. Fig.~\ref{Fig:v2c_overview} illustrates the details of our V2CNet network.

\subsection{Fine-grained Action Classification} We employ a TCN network in the classification branch to encode the temporal information of the video in order to classify the fine-grained actions. Unlike the popular 2D convolution that operates on the 2D feature map of the image, the TCN network performs the convolution across the time axis of the data. Since each video is represented as a sequence of frames and the fine-grained action is the key information for the learning process, using the TCN network to encode this information is a natural choice. Recent work showed that the TCN network can further improve the results in many tasks such as action segmentation and detection~\citep{sun2015_action, Colin17_temporal}.

The input of the classification branch is a set of visual features extracted from the video frames by a pretrained CNN. The output of this branch is a classification score that indicates the most probable action for the input video. In practice, we build a TCN with three convolutional layers to classify the fine-grained actions. After each convolutional layer, a non-linear activation layer (ReLU) is used, followed by a max pooling layer to gradually encode the temporal information of the input video. Here, we notice that instead of performing the max pooling operation in the 2D feature map, the max pooling of TCN network is performed across the time axis of the video. Finally, the output of the third convolutional layer is fed into a fully connected layer with $256$ neurons, then the last layer with only $1$ neuron is used to regress the classification score for the current video.

More formally, given the input feature vector $\mathbf{X}$, three convolutional layers $\mathbf{\Phi}_c$ of the TCN network are defined as follows to encode the temporal information across the time axis of the input demonstration video:

\begin{equation}
\label{eq_tcn_conv} 
\begin{aligned} 
{\mathbf{\Phi}_{c0}}({\mathbf{W}_{c0}},{\mathbf{b}_{c0}}) &= {\mathbf{W}_{c0}}\mathbf{X} + {\mathbf{b}_{c0}}  \\
{\mathbf{\Phi}_{c1}}({\mathbf{W}_{c1}},{\mathbf{b}_{c1}}) &= {\mathbf{W}_{c1}}(\text{MaxPool}(\text{ReLU}({\mathbf{\Phi} _{c0}})) + {\mathbf{b}_{c1}}  \\
{\mathbf{\Phi}_{c2}}({\mathbf{W}_{c2}},{\mathbf{b}_{c2}}) &= {\mathbf{W}_{c2}}(\text{MaxPool}(\text{ReLU}({\mathbf{\Phi} _{c1}})) + {\mathbf{b}_{c2}}  \\
\end{aligned}
\end{equation}
then the third convolutional layer ${\mathbf{\Phi}_{c2}}$ is fed into a fully connected layer $\mathbf{\Phi} _{f0}$ as follows:

\begin{equation}
\label{eq_tcn_fc} 
\begin{aligned} 
{\mathbf{\Phi} _{f0}} ({\mathbf{W}_{f0}},{\mathbf{b}_{f0}}) &= {\mathbf{W}_{f0}}({\text{ReLU} (\mathbf{\Phi} _{c2}})) + {\mathbf{b}_{f0}} \\
{\mathbf{\Phi} _{f1}} ({\mathbf{W}_{f1}},{\mathbf{b}_{f1}}) &= {\mathbf{W}_{f1}}{\mathbf{\Phi} _{f0}} + {\mathbf{b}_{f1}}  \\
\end{aligned}
\end{equation}
where $\mathbf{W}_c$, $\mathbf{b}_c$ and  $\mathbf{W}_f$, $\mathbf{b}_f$ are the weight and bias of the convolutional and fully connected layers. In practice, the filter parameters of three convolutional layers $\mathbf{\Phi}_{c0}$, $\mathbf{\Phi}_{c1}$, $\mathbf{\Phi}_{c2}$ are empirically set to $2048$, $1024$, and $512$, respectively. Note that, the ReLU activation is used in the first fully-connected $\mathbf{\Phi} _{f0}$ layer, while there is no activation in the last fully connected layer $\mathbf{\Phi}_{f1}$ since we want this layer outputs a probability of the classification score for each fine-grained action class.

\subsection{Command Generation} In parallel with the classification branch, we build a translation branch to generate the command sentence from the input video. The architecture of our translation branch is identical to the one we used in the S2SNet.  However, since the translation branch is jointly trained with classification branch, it is encouraged to output the correct fine-grained action as learned by the classification branch. We experimentally show that by simultaneously training both branches, the translation accuracy is significantly improved over the state of the art.

\subsection{Multi-Task Loss}
We train the V2CNet using a joint loss function for both the classification and translation branches as follows:
\begin{equation}
L = L_{cls} + L_{trans}
\end{equation}
where the $L_{cls}$ loss is defined in the classification branch for fine-grained action classification, and $L_{trans}$ is defined in the RNN branch for command generation as in the S2S architecture.

Specially, $L_{cls}$ is the sigmoid cross entropy loss over the groundtruth action classes $C$, and is defined as follows:
\begin{equation}
{L_{cls}} =  - \sum\limits_{i = 1}^C {{\mathbf{y}_i^a}\log ({\hat{\mathbf{y}}_i^a})}
\end{equation}
where $\mathbf{y}^a$ is the groundtruth action label of the current input video, and $\mathbf{\hat y}^a$ is the predicted action output of the classification branch of the network.

\subsection{Training and Inference}
During the training phase, the classification branch uses the visual features to learn the fine-grained actions using the TCN network. In parallel with the classification branch, the translation branch also receives the input via its first LSTM/GRU layer. In particular, at each time step $t$, the input feature $\mathbf{x}_t$ is fed to an LSTM/GRU cell in the first LSTM/GRU layer along with the previous hidden state $\mathbf{h}_{t-1}^e$ to produce the current hidden state $\mathbf{h}_t^e$. After all the input features are exhausted, the word embedding and the hidden states of the first LSTM/GRU layer are fed to the second LSTM/GRU layer. This layer converts the inputs into a sequence of words by maximizing the log-likelihood of the predicted word. This decoding process is performed sequentially for each word until the network generates the \textsf{EOC} token. Since both the classification and translation branches are jointly trained using a single loss function, the weight and bias parameters of both branches are updated using the same gradient signal through backpropagation. 

During the inference phase, the input for the network is only the visual features of the testing video. The classification branch uses these visual features to generate the probabilities for all action classes. The final action class is chosen from the class with the highest classification score. Similarly, the visual features are fed into two LSTM/GRU layers to sequentially generate the output words as the command. Note that, unlike the training process, during inference the input for the second LSTM/GRU layer in the translation branch is only the hidden state of the first LSTM/GRU layer. The final command sentence is composed of the first generated word and the last word before the \textsf{EOC} token.

\section{Experiments} \label{Sec:_act_exp}
\subsection{Dataset}

Recently, many datasets have been proposed in the video captioning field~\citep{Xu16_MSR_Dataset}. However, these datasets only provide general descriptions of the video and there is no detailed understanding of the action. The groundtruth captions are also written using natural language sentences which can not be used directly in robotic applications. Motivated by these limitations, we introduce a new \textit{video-to-command} (IIT-V2C) dataset which focuses on \textit{fine-grained} action understanding~\citep{lea2016learning}. Our goal is to create a new large-scale dataset that provides fine-grained understanding of human actions in a grammar-free format. This is more suitable for robotic applications and can be used with deep learning methods.

\begin{figure}[ht]
\centering
\footnotesize

  \stackunder[2pt]{\includegraphics[width=0.99\linewidth, height=0.18\linewidth]{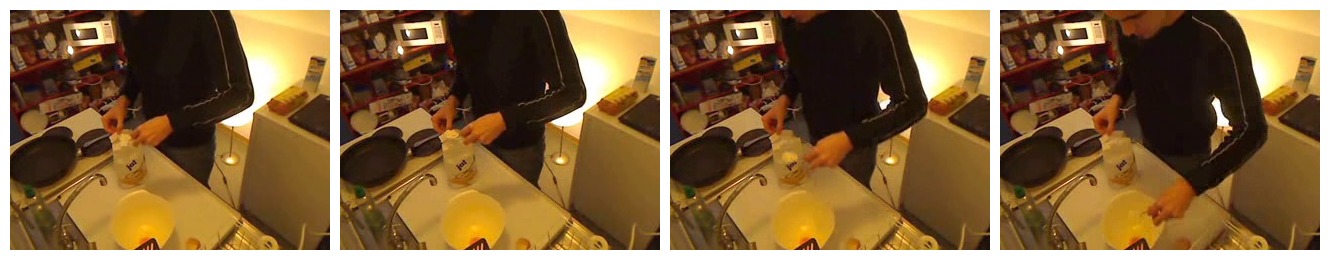}}  				
  				  {\datasetCaption {lefthand transfer powder to bowl} {transferring}}
  				  %\vspace{2ex} 
  %\hspace{2ex}%
  \stackunder[2pt]{\includegraphics[width=0.99\linewidth, height=0.18\linewidth]{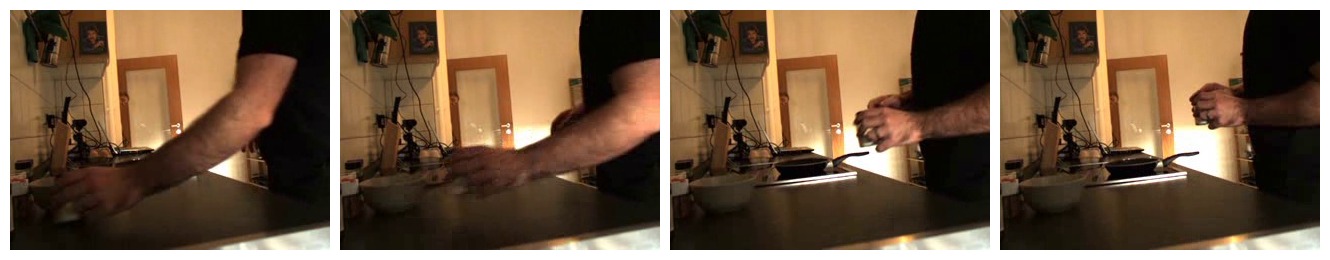}}  
  				  {\datasetCaption {lefthand carry salt box} {carrying}}

  \stackunder[2pt]{\includegraphics[width=0.99\linewidth, height=0.18\linewidth]{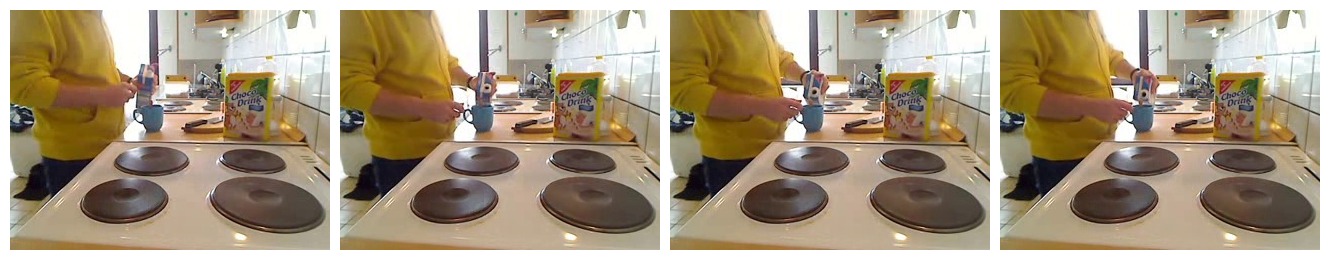}}  
    			  {\datasetCaption {righthand pour milk} {pouring} }
  				  %\vspace{2ex}
  
  \stackunder[2pt]{\includegraphics[width=0.99\linewidth, height=0.18\linewidth]{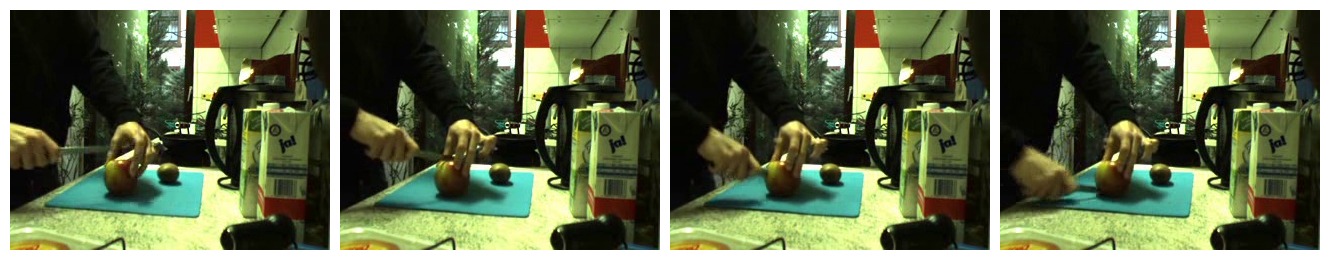}}  
    			  {\datasetCaption {bothhand cut apple} {cutting} }
  				  %\vspace{0ex}
  				  
  %\hspace{-0.25cm}%
 %\hspace{-0.25cm}%
\vspace{-0.0ex}
\caption{Example of human demonstration clips and labeled groundtruths for the command sentence and fine-grained action class in our IIT-V2C dataset. The clips were recorded from challenging viewpoints with different lighting conditions.}
\label{fig_exp_figure} 
\end{figure}

\begin{figure*}[ht]
  \centering
 	\includegraphics[scale=0.32]{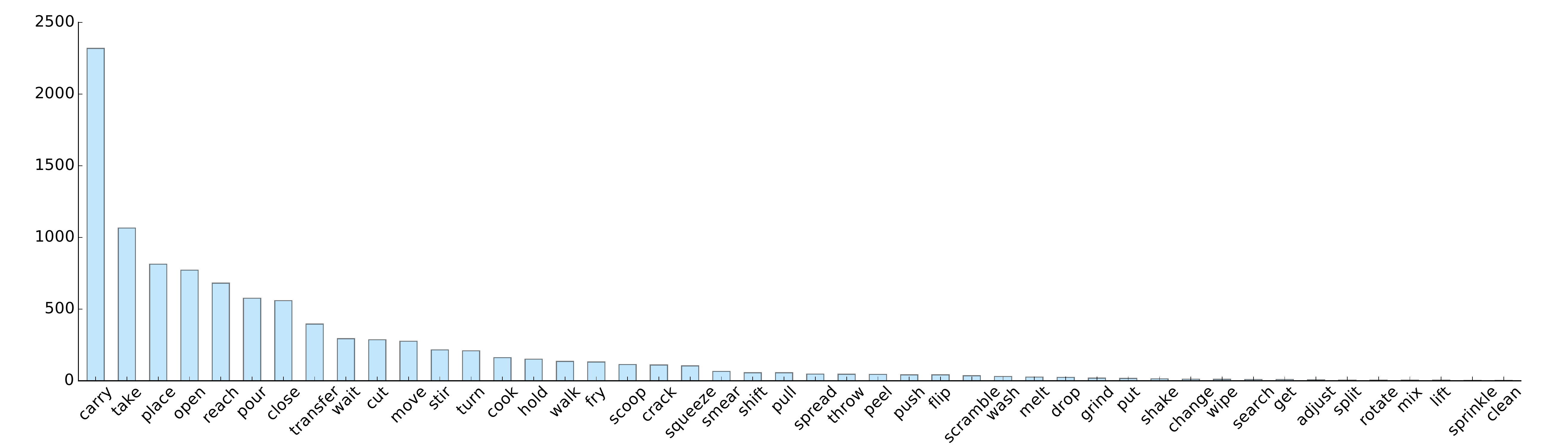}
 \vspace{-1.0ex}
 \caption{The distribution of the fine-grained action classes in our IIT-AFF dataset.}
 \label{fig_act_distribution}
\end{figure*}

\textbf{Video annotation} 
Since our main purpose in this work is to develop a framework that can be used by real robots for manipulation tasks, we are only interested in the videos that have human demonstrations. To this end, the raw videos in the Breakfast dataset~\citep{Kuehne14_BF_Dataset} are best suited to our purpose since they were originally designed for activity recognition. We only reuse the raw videos from the Breakfast dataset and manually segment each video into short clips in a fine granularity level. Each short clip is then annotated with a \textit{command sentence} in grammar-free that describes the current human action. For each command sentence, we use the Stanford POS Tagger~\citep{Toutanova_tagger} to automatically extract the verb from the command. This extracted verb is then considered as the fine-grained \textit{action groundtruth} for the current video.

%//////////////////////////////////////////////
\newcolumntype{g}{>{\columncolor{Gray}}c}
\begin{table*}[!ht]
\centering\ra{1.2}
\caption{The translation results on IIT-V2C dataset.}
\renewcommand\tabcolsep{1.5pt}
\label{tb_result_v2c}
\hspace{2ex}

\begin{tabular}{@{}r|c|c|cccccccc@{}}  %%% original white
%\begin{tabular}{@{}r|c|c|gggggggg@{}}   %%% gray color
\toprule 					 &
\small RNN  	& 
\small Feature  & 

\small Bleu\_1  & 
\small Bleu\_2  & 
%\multirow{1}{*}[2.5pt]{\scriptsize DeepLab~\cite{Chen2016_deeplab}} & 
\small Bleu\_3  &
\small Bleu\_4  & 
%\mu\ltirow{1}{*}[2.5pt]{\scriptsize BB-CNN} & 
\small METEOR  &
\small ROUGE\_L &
\small CIDEr \\

\midrule

S2VT						&\scriptsize{LSTM} & \scriptsize{VGG16, AlexNet}		& 0.383   & 0.265   & 0.201	& 0.159	& 0.183    & 0.382   & 1.431     \\
S2VT 						&\scriptsize{LSTM} & \scriptsize{ResNet50, AlexNet}	& 0.397   & 0.280   & 0.219	& 0.177	& 0.196    & 0.401   & 1.560     \\
SGC							&\scriptsize{LSTM} & \scriptsize{Inception}   	& 0.370   & 0.256   & 0.198	& 0.161	& 0.179    & 0.371   & 1.422     \\
SCN							&\scriptsize{LSTM} & \scriptsize{ResNet50, C3D}  	& 0.398   & 0.281   & 0.219	& 0.190	& 0.195    & 0.399   & 1.561     \\

\cline{1-10}
\multirow{6}{*}{S2SNet} & \multirow{3}{*}{\scriptsize{LSTM}} 	&\scriptsize{VGG16} 		& 0.372   & 0.255  	& 0.193	& 0.159	& 0.180    & 0.375   & 1.395     \\
															  &	&\scriptsize{Inception}	& 0.400   & 0.286   & 0.221	& 0.178	& 0.194    & 0.402   & 1.594     \\   
												 			  &	&\scriptsize{ResNet50}	& 0.398   & 0.279   & 0.215	& 0.174	& 0.193    & 0.398   & 1.550     \\
									  \cline{2-10}
									  & \multirow{3}{*}{\scriptsize{GRU}}
																&\scriptsize{VGG16} 		& 0.350   & 0.233   & 0.173	& 0.137	& 0.168    & 0.351   & 1.255     \\
														 	  &	&\scriptsize{Inception}	& 0.391   & 0.281  	& 0.222	& 0.188	& 0.190    & 0.398   & 1.588     \\
													 		  &	&\scriptsize{ResNet50}	& 0.398   & 0.284   & 0.220	& 0.183	& 0.193    & 0.399   & 1.567     \\
\cline{1-10}

\multirow{6}{*}{V2CNet} & \multirow{3}{*}{\scriptsize{LSTM}}	& \scriptsize{VGG16} 	& 0.391   & 0.275   & 0.212 & 0.174 & 0.189    & 0.393   & 1.528     \\
					   									  	&   & \scriptsize{Inception} & 0.401   & 0.289   & 0.227 & 0.190 & 0.196    & 0.403   & 1.643     \\
					   									  	&   & \scriptsize{ResNet50} 	&\textbf{0.406}   & \textbf{0.293}   & \textbf{0.233} & \textbf{0.199} & \textbf{0.198}    & \textbf{0.408}   & \textbf{1.656}	 \\
					   			   \cline{2-10}

						& \multirow{3}{*}{\scriptsize{GRU}} 	& \scriptsize{VGG16} 	& 0.389   & 0.267   & 0.208 & 0.172 & 0.186    & 0.387   & 1.462     \\
					  									  	&   & \scriptsize{Inception} & 0.402   & 0.285   & 0.224 & 0.189 & 0.196    & 0.405   & 1.618     \\
					  									  	&   & \scriptsize{ResNet50}  & 0.403   & 0.288   & 0.226 & 0.191 & 0.196    & 0.403   & 1.596     \\

\bottomrule
\end{tabular}
\end{table*}

\textbf{Dataset statistics} Overall, we reuse $419$ videos from the Breakfast dataset. These videos were captured when humans performed cooking tasks in different kitchens, then encoded with $15$ frames per second. The resolution of demonstration videos is $320 \times 240$. We segment each video (approximately $2-3$ minutes long) into around $10-50$ short clips (approximately $1-15$ seconds long), resulting in $11,000$ unique short videos. Each short video has a single command sentence that describes human actions. From the groundtruth command sentence, we extract the verb as the action class for each video, resulting in an action set with $46$ classes (e.g., cutting, pouring, etc.). Fig.~\ref{fig_exp_figure} shows some example frames of groudtruth command sentences and action classes in our new dataset. In Fig.~\ref{fig_act_distribution}, the distribution of the fine-grained action classes is also presented. Although our new-form dataset is characterized by its grammar-free property for the convenience in robotic applications, it can easily be adapted to classical video captioning task by adding the full natural sentences as the new groundtruth for each video.

\subsection{Evaluation and Baseline}
\textbf{Evaluation Metric} We use the standard evaluation metrics in the video captioning field~\citep{Xu16_MSR_Dataset} (Bleu, METEOR, ROUGE-L, and CIDEr) to report the translation results of our V2CNet. By using the same evaluation metrics, we can directly compare our results with the recent state of the art in the video captioning field.

\textbf{Baseline} The translation results of our V2CNet are compared with the following state of the art in the field of video captioning: S2VT~\citep{Venugopalan2016}, SGC~\citep{Ramanishka2017cvpr}, and SCN~\citep{SCN_CVPR2017}. In S2VT, the authors used the encoder-decoder architecture with LSTM to encode the visual features from RGB images (extracted by ResNet50 or VGG16) and optical flow images (extracted by AlexNet~\citep{Alex12}). In SGC, the authors also used the encoder-decoder architecture and LSTM, however, this work integrated a saliency guided method as the visual attention mechanism, while the visual features are from the Inception network. The authors in SCN~\citep{SCN_CVPR2017} first combined the visual features from ResNet and temporal features from C3D~\citep{Tran15_C3D} network, then extracted semantic concepts (i.e., tags) from the video frames. All these features were learned in a LSTM as a semantic recurrent neural network. Finally, we also compare the V2CNet results with our early work (EDNet~\citep{Nguyen_V2C_ICRA18}). The key difference between EDNet and V2CNet is the EDNet does not use the TCN network to jointly train the fine-grained action classification branch and the translation branch as in the V2CNet. For all methods, we use the code provided by the authors of the associated papers for the fair comparison.

\textbf{V2CNet Implementation} We use $512$ hidden units in both LSTM and GRU in our implementation. The first hidden state of LSTM/GRU is initialized uniformly in $[-0.1, 0.1]$. We set the number of frames for each input video at $n=30$. Subsequently, we consider each command has maximum $30$ words. If there are not enough $30$ frames/words in the input video/command, we pad the mean frame/empty word at the end of the list until it reaches $30$. The mean frame is composed of pixels with the same mean RGB value from the ImageNet dataset (i.e., $(104, 117, 124)$). We use $70\%$ of the IIT-V2C dataset for training and the remaining $30\%$ for testing. During the training phase, we only accumulate the softmax losses of the real words to the total loss, while the losses from the empty words are ignored. We train all the variations of V2CNet for $300$ epochs using Adam optimizer~\citep{kingma2014adam} with a learning rate of $0.0001$, and the batch size is empirically set to $16$. The training time for each variation of the V2CNet is around $8$ hours on an NVIDIA Titan X GPU.

\subsection{Translation Results}

Table~\ref{tb_result_v2c} summarizes the translation results on the IIT-V2C dataset. This table clearly shows that our V2CNet with ResNet50 feature achieves the highest performance in all metrics: Blue\_1, Blue\_2, Blue\_3, Blue\_4, METEOR, ROUGE\_L, and CIDEr. Our proposed V2CNet also outperforms recent the state-of-the-art methods in the video captioning field (S2VT, SGC, and SCN) by a substantial margin. In particular, the best CIDEr score of V2CNet is $1.656$, while the CIDEr score of the closest runner-up SCN is only $1.561$. Furthermore, compared with the results by EDNet, V2CNet also shows a significant improvement in all experiments when different RNN types (i.e., LSTM or GRU) or visual features (i.e., VGG16, Inception, ResNet50) are used. These results demonstrate that by jointly learning both the fine-grained actions and the commands, the V2CNet can effectively encode both the visual features to generate the commands while is able to understand the fine-grained action from the demonstration videos. Therefore, the translation results are significantly improved.

Overall, we have observed a consistent improvement of our proposed V2CNet over EDNet in all variation setups. The improvement is most significant when the VGG16 feature is used. In particular, while the results of EDNet using VGG16 feature with both LSTM and GRU networks are relatively low, our V2CNet shows a clear improvement in these cases. For example, the Blue\_1 scores of EDNet using VGG16 feature are only $0.372$ and $0.350$ with the LSTM and GRU network respectively, while with the same setup, the V2CNet results are $0.391$ and $0.389$. We also notice that since EDNet only focuses on the translation process, it shows a substantial gap in the results between the VGG16 feature and two other features. However, this gap is gradually reduced in the V2CNet results. This demonstrates that the fine-grained action information in the video plays an important role in the task of translating videos to commands, and the overall performance can be further improved by learning the fine-grained actions.

From this experiment, we notice that there are three main factors that affect the results: the translation architecture, the input visual feature, and the external mechanism such as visual or action attention. While the encoder-decoder architecture is widely used to interpret videos to sentences, recent works focus on exploring the use of robust features and attention mechanism to improve the results. Since the IIT-V2C dataset contains mainly the fine-grained human demonstrations, the visual attention mechanism (such as in SGC architecture) does not perform well as in the normal video captioning task. On the other hand, the action attention mechanism and the input visual features strongly affect the final results. Our experiments show that the use of TCN network as the action attention mechanism clearly improves the translation results. In general, we note that the temporal information of the video plays an important role in this tasks. By extracting this information offline (e.g., from optical flow images as in S2VT, or with C3D network as in SCN), or learning it online as in our V2CNet, the translation results can be further improved.

\begin{figure*}
\centering
\footnotesize
 %\stackunder[5pt]{\includegraphics[width=0.49\linewidth, height=0.09\linewidth]{figures/4_exp/result.pdf}}
  \stackunder[2pt]{\includegraphics[width=0.99\linewidth, height=0.17\linewidth]{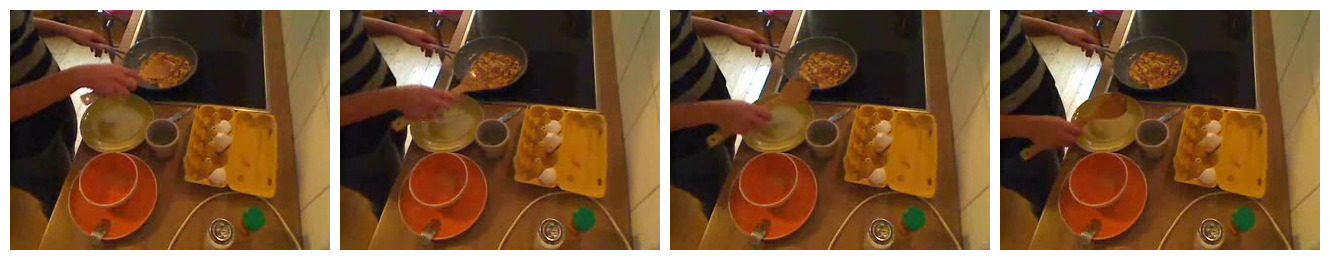}}  				
  				  {\tableCaption {righthand carry spatula} {righthand carry spatula} {lefthand reach stove} {lefthand reach pan} {righthand carry spatula} {righthand take spatula} }
  \hspace{0.25cm}%
  
  \stackunder[2pt]{\includegraphics[width=0.99\linewidth, height=0.17\linewidth]{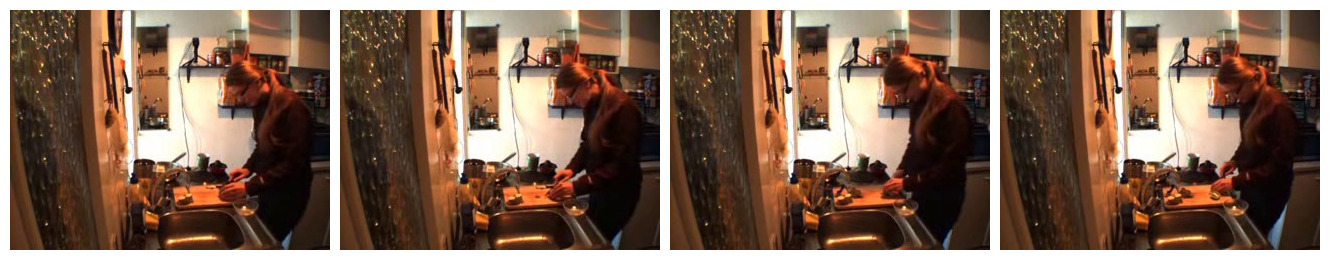}}  
  				  {\tableCaption {righthand cut fruit} {righthand cut fruit} {righthand cut fruit} {righthand cut fruit} {righthand cut fruit} {righthand cut fruit} }

 %\stackunder[5pt]{\includegraphics[width=0.99\linewidth, height=0.09\linewidth]{figures/4_exp/result.pdf}} 
  \stackunder[2pt]{\includegraphics[width=0.99\linewidth, height=0.17\linewidth]{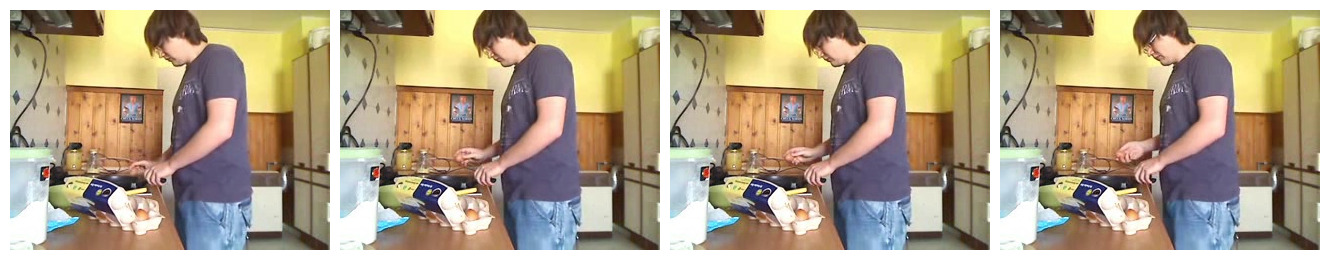}}  
    			  {\tableCaption {righthand crack egg} {righthand carry egg} {lefthand reach spatula} {righthand carry egg} {righthand crack egg} {lefthand carry egg} }
  \hspace{0.25cm}%
  
 %\stackunder[5pt]{\includegraphics[width=0.49\linewidth, height=0.09\linewidth]{figures/4_exp/result.pdf}}  
  \stackunder[2pt]{\includegraphics[width=0.99\linewidth, height=0.17\linewidth]{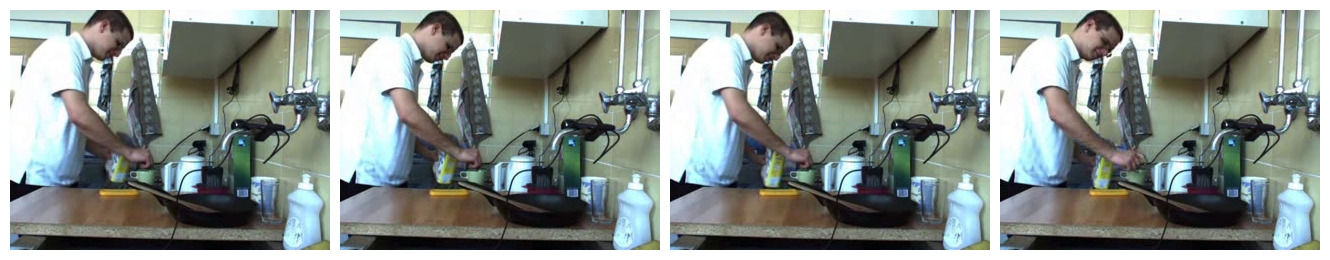}}    
				  {\tableCaption {righthand stir milk} {righthand hold teabag} {righthand place kettle} {righthand take cacao}{righthand stir coffee} {righthand stir milk} }
\hspace{0.25cm}% 
  				  
%  \stackunder[2pt]{\includegraphics[width=0.99\linewidth, height=0.17\linewidth]{chapter4_action/figures/4_exp/p44_milk_all.jpg}}  
%    			  {\tableCaption {righthand transfer cacao to cup} {righthand reach cup} {righthand hold pan} {righthand pour cacao} {righthand transfer cacao to cup}{righthand transfer fruit} }
%  				  \vspace{0ex}
  
  \stackunder[2pt]{\includegraphics[width=0.99\linewidth, height=0.17\linewidth]{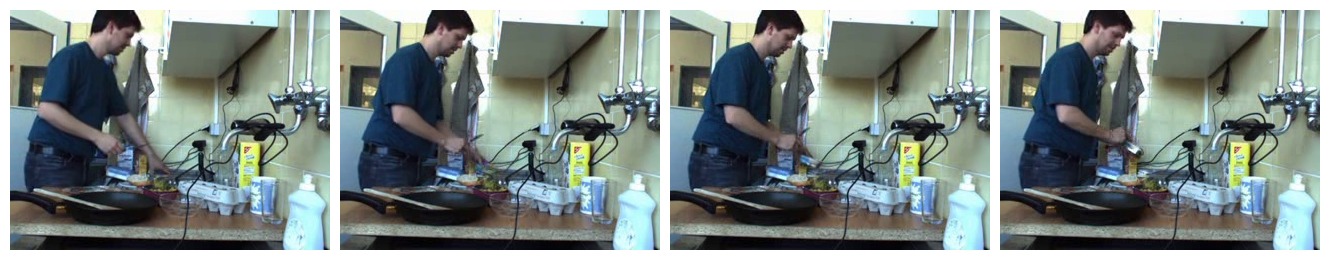}}    
				  {\tableCaption {lefthand take butter box} {lefthand reach teabag} {righthand place kettle} {righthand take kettle} {lefthand take topping}{lefthand reach milk bottle} }

  %\hspace{-0.25cm}%
 %\hspace{-0.25cm}%
\vspace{1ex}
\caption{Example of translation results of our S2SNet, V2CNet, and other methods in comparison with the groundtruth (\textbf{GT}).}

\label{Fig:main_result} 
\end{figure*}

Fig.~\ref{Fig:main_result} shows some comparisons between the commands generated by our V2CNet and other methods, compared with the groundtruth on the test videos of the IIT-V2C dataset. In general, our V2CNet is able to generate good predictions in many examples, while the results of other methods are more variable. In comparison with EDNet, V2CNet shows more generated commands with the correct fine-grained actions. We notice that in addition to the generated commands that are identical with the groundtruth, many other output commands are relevant. These qualitative results show that our V2CNet can further improve the translation results. However, there are still many wrong predictions in the results of all the methods. Since the IIT-V2C dataset contains the fine-grained actions, while the visual information is also difficult (e.g, the relevant objects are small and usually are covered by the hand, etc.). This makes the problem of translating videos to commands is more challenging than the normal video captioning task since the network has to rely on the minimal information to predict the command sentence.

To conclude, the extensive experiments using different feature extraction methods and RNNs show that our V2CNet successfully encodes the visual features and generates the associated command for each video. Our proposed method also outperforms recent state of the art by a substantial margin. The key idea that improves the translation results is the integration of the TCN network into the traditional encoder-decoder architecture to effectively learn the fine-grained actions in the video.

\subsection{Ablation Studies}

Although the traditional captioning metrics such as Bleu, METEOR, ROUGE\_L, and CIDEr give us the quantitative evaluation about the translation results, they use all the words in the generated commands for the evaluation, hence do not provide details about the accuracy of the fine-grained human actions. To analyze the prediction accuracy of the fine-grained actions, we conduct the following experiments: For both EDNet and V2CNet, we use the LSTM network and the visual features from the VGG16, Inception, and ResNet to generate both the commands from the translation branch, and the action class from the classification branch. The predicted actions of the translation branch are then automatically extracted by using the Stanford POS Tagger~\citep{Toutanova_tagger} to select the verb from the generated commands, while the classification branch gives us directly the fine-grained action class. For each variation of the networks, we report the success rate of the predicted output as the percentage of the correct predictions over all the testing clips. Intuitively, this experiment evaluates the accuracy of the fine-grained action prediction when we consider the translation branch also outputs the fine-grained action classes.

%//////////////////////////////////////////////
\newcolumntype{g}{>{\columncolor{Gray}}c}
\begin{table}[!ht]
\centering\ra{1.2}
\caption{The fine-grained action classification success rate.}
\renewcommand\tabcolsep{3.5pt}
\label{tb_result_action}
\hspace{2ex}

\begin{tabular}{@{}r|c|c@{}}  %%% original white
%\begin{tabular}{@{}r|c|c|gggggggg@{}}   %%% gray color
\toprule 					 &
\small Feature 	& 
\small Success Rate  \\

\midrule
\multirow{3}{*}{S2SNet~\citep{Nguyen_V2C_ICRA18}}  	&VGG16 		& 30.17\%     \\
												    &Inception	& 32.88\%     \\   
												    &ResNet50	& 32.71\%     \\									  
\cline{1-3}

\multirow{3}{*}{\makecell{V2CNet \\(translation branch)}} 	& VGG16 	& 31.75\%   \\
					   										& Inception & 34.41\%   \\
					   		    							& ResNet50 	& \textbf{34.81}\%   \\
\cline{1-3}					   			  
\multirow{3}{*}{\makecell{V2CNet \\(classification branch)}}& VGG16 	& 31.94\%   \\
					   										& Inception & 34.52\%   \\
					   		    							& ResNet50 	& 34.69\%   \\

\bottomrule
\end{tabular}
\end{table}

Table~\ref{tb_result_action} summaries the success rate of the fine-grained action classification results. Overall, we observe a consistent improvement of V2CNet over EDNet in all experiments using different visual features. In particular, when the actions are extracted from the generated commands of the translation branch in V2CNet, we achieve the highest success rate of $34.81\%$ with the ResNet50 features. This is a $2.1\%$ improvement over EDNet with ResNet50 features, and $4.64\%$ over EDNet with VGG16 features. While Table~\ref{tb_result_action} shows that the V2CNet clearly outperforms EDNet in all variation setups, the results of the translation branch and classification branch of V2CNet are a tie. These results demonstrate that the use of the TCN network in the classification branch is necessary for improving both the translation and classification accuracy. However, the overall classification success rate is relatively low and the problem of translating videos to commands still remains very challenging since it requires the understanding of the fine-grained actions. 

\subsection{Robotic Applications}

Similar to~\citep{Yang2015}, our long-term goal is to develop a framework that allows the robot to perform various manipulation tasks by just ``\textit{watching}" the input video. While the V2CNet is able to interpret a demonstration video to a command, the robot still needs more information such as scene understanding (e.g., object affordance, grasping frame, etc.), and trajectory planner to complete a manipulation task. In this work, we build a framework based on three basic components: action understanding, affordance detection, and trajectory generation. In practice, the proposed V2CNet is first used to let the robot understand the human demonstration from a video, then the AffordanceNet~\citep{AffordanceNet17} which is trained on IIT-AFF dataset~\citep{Nguyen2017_Aff} is used to localize the object affordances and the grasping frames. Finally, the motion planner is used to generate the trajectory for the robot in order to complete the manipulation tasks.

%\begin{figure*}[ht]
%  \centering
%% \subfigure[Human demonstration]{\label{fig_rba}\includegraphics[width=0.99\linewidth, height=0.3\linewidth]{figures/5_robot/pour_human.pdf}}
%% \subfigure[WALK-MAN replication]{\label{fig_rbb}\includegraphics[width=0.99\linewidth, height=0.3\linewidth]{figures/5_robot/pour_walkman.pdf}}
%% \subfigure[WALK-MAN replication]{\label{fig_rbc}\includegraphics[width=0.99\linewidth, height=0.3\linewidth]{figures/5_robot/pour_walkman.pdf}}
% \subfigure[Human demonstration]{\label{fig_rba}\includegraphics[width=0.99\linewidth, height=0.16\linewidth]{figures/5_robot/pour_human_all.pdf}}
% \subfigure[WALK-MAN replication]{\label{fig_rbb}\includegraphics[width=0.99\linewidth, height=0.16\linewidth]{figures/5_robot/pour_walkman_all.pdf}}
% %\subfigure[WALK-MAN replication]{\label{fig_rbc}\includegraphics[width=0.99\linewidth, height=0.16\linewidth]{figures/5_robot/pour_walkman_all.pdf}}
% 
% \vspace{1.0ex}
% \caption{Example of manipulation tasks performed by WALK-MAN using our proposed framework. \textbf{(Left Side)} Pick and place task. \textbf{(Right Side)} Pouring task. More illustrations can be found in the supplemental video.}
% \label{fig_robot_imitation}
%\end{figure*}

\begin{figure*}[ht]
  \centering
 \subfigure[Pick and place task using WALK-MAN]{\label{fig_resize_map_a}\includegraphics[width=0.99\linewidth, height=0.165\linewidth]{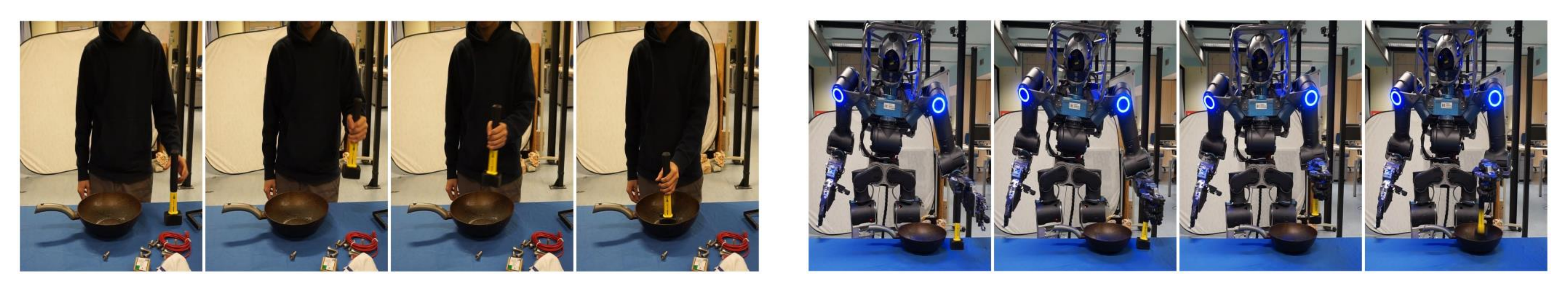}}
 \subfigure[Pouring task using WALK-MAN]{\label{fig_resize_map_b}\includegraphics[width=0.99\linewidth, height=0.165\linewidth]{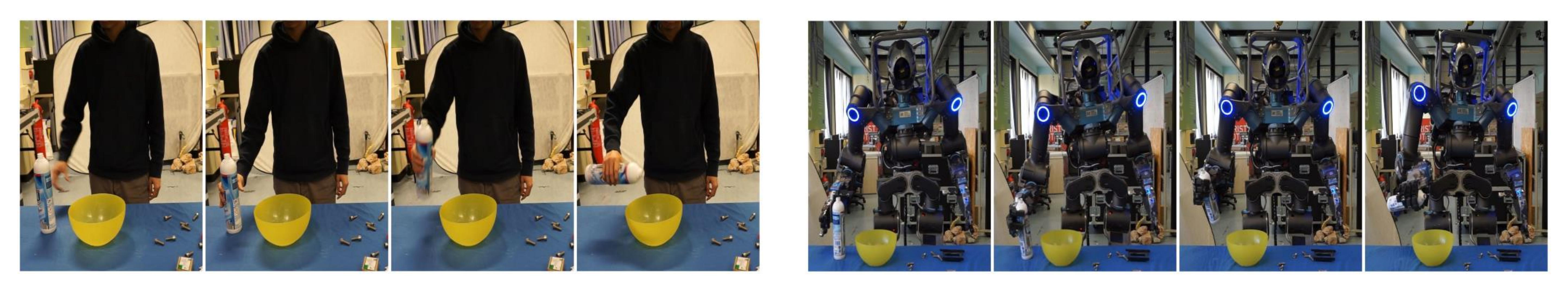}}
 \subfigure[Pick and place task using UR5 arm]{\label{fig_resize_map_b}\includegraphics[width=0.99\linewidth, height=0.165\linewidth]{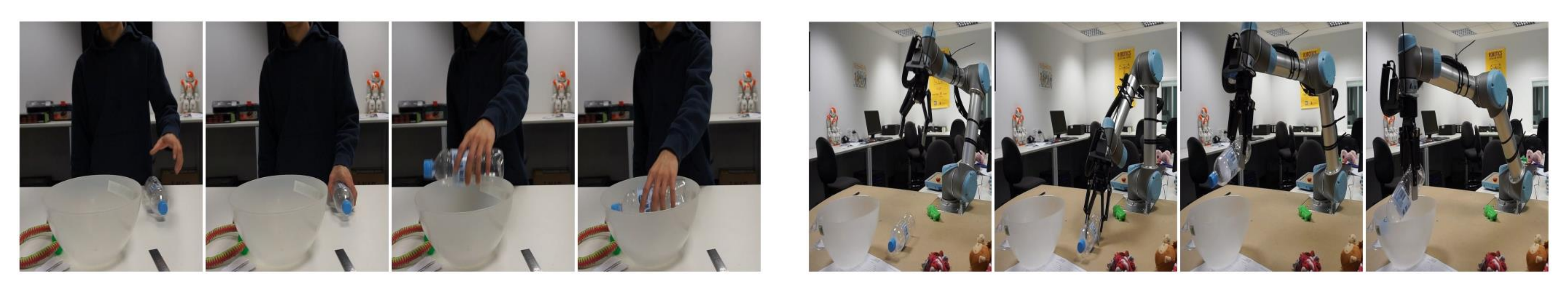}}

 \vspace{2.0ex}
 \caption{Example of manipulation tasks performed by WALK-MAN and UR5 arm using our proposed framework: \textbf{(a)} Pick and place task using WALK-MAN, \textbf{(b)} Pouring task using WALK-MAN, and \textbf{(c)} Pick and place task using UR5 arm. The frames from human instruction videos are on the left side, while the robot performs actions on the right side. We notice that there are two sub-tasks (i.e., two commands) in these tasks: grasping the object and manipulating it. More illustrations can be found in the supplemental video.}
 \label{fig_robot_imitation}
\end{figure*}

We conduct the experiments using two robotic platforms: WALK-MAN~\citep{Niko2016_full} humanoid robot and UR5 arm. WALK-MAN is a full-size humanoid robot with 31 DoF and two underactuated hands. Each hand has five fingers but only 1 DoF, and is controlled by one single motor. Therefore, it can only simultaneously generate the open-close grasping motion of all five fingers. We use XBotCore software architecture~\citep{muratore2017xbotcore} to handle the communication with the robot, while the full-body motion is planned by OpenSoT library~\citep{Rocchi15}. The vision system of WALK-MAN is equipped with a Multisense SL camera, while in UR5 arm we use a RealSense camera to provide visual data for the robot. Due to the hardware limitation of the end-effector in both robots, we only demonstrate the tasks that do not require precise skills of the fingers (e.g., cutting, hammering), and assume that the motion planning library can provide feasible solutions during the execution.

For each task presented by a demonstration video, V2CNet first generates a command sentence, then based on this command the robot uses its vision system to find relevant objects and plan the actions. Fig.~\ref{fig_robot_imitation} shows some tasks performed by WALK-MAN and UR5 arm using our framework. For a simple task such as ``righthand grasp bottle", the robots can effectively repeat the human action through the command. Since the output of our translation module is in grammar-free format, we can directly map each word in the command sentence to the real robot command. In this way, we avoid using other modules as in~\citep{Tellex2011} to parse the natural command into the one that uses in the real robot. The visual system also plays an important role in our framework since it provides the target frames (e.g., grasping frame, ending frame) for the planner. Using our approach, the robots can also complete long manipulation tasks by stacking a list of demonstration videos in order for the translation module. Note that, for the long manipulation tasks, we assume that the ending state of one task will be the starting state of the next task. Overall, the robots successfully perform various tasks such as grasping, pick and place, or pouring. Our experimental video is available at: \url{ https://sites.google.com/site/v2cnetwork}.

%\begin{center}
%\vspace{-1.0ex}
%\url{ https://sites.google.com/site/v2cnetwork}
%\end{center}

%\input{chapter4_action/4_exp}
%\input{chapter4_action/5_discussion}
%\input{chapter4_action/6_conclusions}

\chapter{Conclusion}
\label{ch_conclusion}
Despite recent rapid progress in computer vision and deep learning, it is clear that many challenges still remain in the field of robotic vision. In this work, we proposed different approaches for two fundamental robotic vision problems: affordance detection and fine-grained action understanding. The experimental results on publicly available datasets show that our methods not only achieve state-of-the-art performance but also can be used in various manipulation tasks. 

Currently, our AffordanceNet architecture provides a good solution to detect object affordances in both real and simulation images. It processes the input image at $150ms$ on a modern GPU, and well suitable for manipulation applications with the real robot or in simulation. However, the main limitation of AffordanceNet is it still requires a huge amount of memory during the training and testing phases. This is because of the use of a series of deconvolutional layers in our architecture. Therefore, an interesting future research direction on this task is to develop a new compact architecture that can maintain the accuracy while uses less memory.

In this work, we divide the robot imitation task into two steps: the understanding step and the imitation step. We form the understanding step as a fine-grained video captioning task and solve it as a visual translation problem. From the extensive experiments on the IIT-V2C dataset, we have observed that the translation accuracy not only depends on the visual information of each frame but also depends on the temporal information across the video. By using the TCN network to encode the temporal information, we achieved a significant improvement over the state of the art. Despite this improvement, we acknowledge that this task remains very challenging since it requires the fine-grained understanding of human actions, which is still an unsolved problem in computer vision~\citep{lea2016learning}.

From a robotic point of view, by explicitly representing the human demonstration as a command then combining it with the vision and planning module, we do not have to handle the domain shift problem~\citep{yu2018one} in the real experiment. Another advantage of this approach is we can reuse the strong state-of-the-art results from the vision and planning fields. However, its main drawback is the reliance on the accuracy of each component, which may become the bottleneck in real-world applications. For example, our framework currently relies solely on the vision system to provide the grasping frames and other useful information for the robot. Although recent advances in deep learning allow us to have powerful recognition systems, these methods are still limited by the information presented in the training data, and cannot provide the fully semantic understanding of the scene~\citep{AffordanceNet17}.

From the robotic experiments, we notice that while the vision and planning modules can produce reasonable results, the translation module is still the weakest part of the framework since its results are more variable. Furthermore, we can also integrate the state-of-the-art LfD techniques to the planning module to allow the robots to perform more useful tasks, especially the ones that require precise skill such as ``cutting" or ``hammering". However, in order to successfully perform these tasks, the robots also need to be equipped with sufficient end-effector. From a vision point of view, although our V2CNet can effectively translate videos to commands, the current network architecture considers each demonstration video equally and does not take into account the order of these videos. In real-world scenarios, the order of the actions also plays an important role to complete a task (e.g., ``putting on sock" should happen before ``wearing shoe"). Currently, we assume that the order of the sub-videos in a long manipulation task is known. Therefore, another interesting problem is to develop a network that can simultaneously segment a long video into short clips and generate a command sentence for each clip. 

% body of thesis comes here

\appendix
% appendices come here

\addcontentsline{toc}{chapter}{Bibliography}
%\bibliographystyle{alpha}
%\bibliography{bibliography/bibliography}
\bibliography{0_class/reference}

\end{document}